\documentclass{article}

\PassOptionsToPackage{numbers, compress}{natbib}

\usepackage[english]{babel}

\usepackage[letterpaper,top=2cm,bottom=2cm,left=3cm,right=3cm,marginparwidth=1.75cm]{geometry}
\usepackage[parfill]{parskip}
\usepackage[utf8]{inputenc} 
\usepackage[T1]{fontenc}    
\usepackage{microtype}      

\usepackage{amsmath}        
\usepackage{amsfonts}       
\usepackage{amssymb}
\usepackage{amsthm}
\usepackage{mathtools}
\usepackage{nicefrac}       

\usepackage{graphicx}
\usepackage{booktabs}       
\usepackage{tabularx}       
\usepackage{multirow}
\usepackage[table]{xcolor}  
\usepackage{float}
\usepackage{rotating}       
\usepackage{subcaption}     
\usepackage{wrapfig}
\usepackage{pifont}
\usepackage[shortlabels]{enumitem}
\usepackage{tcolorbox}
\tcbuselibrary{skins,listings,breakable}

\usepackage{url}            
\usepackage{natbib}
\usepackage[colorlinks=true, allcolors=blue]{hyperref}

\newcommand{\icwb}{$\mathsf{ICWBench}$}

\title{Learning to Follow In-Context Watermark Instructions \\ via Self-Distillation}

\author{
\setlength{\tabcolsep}{15pt}
  \begin{tabular}{ccc}
    Yepeng Liu$^{*,\dagger}$ & Tianyi Chen$^{*}$ & Xuandong Zhao \\
    UC Santa Barbara & UC San Diego & UC Berkeley
  \end{tabular} \\[3ex]
  \setlength{\tabcolsep}{18pt}
  \begin{tabular}{cc}
    Dawn Song & Yuheng Bu \\
    UC Berkeley & UC Santa Barbara
  \end{tabular}
}

\date{}

\begin{document}
\maketitle

{\renewcommand{\thefootnote}{\fnsymbol{footnote}}%
\footnotetext[1]{These authors contributed equally to this work.}%
\footnotetext[2]{Correspondence to: yepengliu@ucsb.edu}}

\begin{abstract}
In-context watermarking (ICW) prepends an instruction to a query asking the model to embed a statistically detectable signal in its response. It thus equips LLMs with a watermarking interface that third parties can invoke without access to model internals. Its reliability hinges on the LLM following the instruction without degrading answer quality, yet how well current LLMs do so has not been measured. We introduce $\mathsf{ICWBench}$, a benchmark of three verifiable ICW instruction families, each scored on both detectability and answer quality. Evaluating 14 frontier proprietary and open-source LLMs, we find that none of the evaluated LLMs achieves both objectives across all three families. To address this, we propose a self-contained two-stage training method, requiring no distillation from a stronger model, no manual annotation, and no pre-existing ICW IF ability. The first stage, self-distillation with logits perturbation (SDLP), uses the same base LLM as both teacher and student: an instruction-equivalent decoding-time logits perturbation makes the teacher follow the ICW instruction, and the student is trained to match the teacher's output distribution. The second stage applies reinforcement learning with the automatic verifier as the reward. Applied to Qwen3-14B and GPT-OSS-20B, our method raises average TPR@$1\%$FPR across three ICW instructions from $0.100$ to $0.974$ and from $0.337$ to $0.968$, respectively, while maintaining high response quality under both perplexity evaluation and LLM-as-a-Judge. Our code is available at \url{https://github.com/yepengliu/ICW-IF}.

\begin{figure}[h]
\centering
\includegraphics[width=0.8\linewidth]{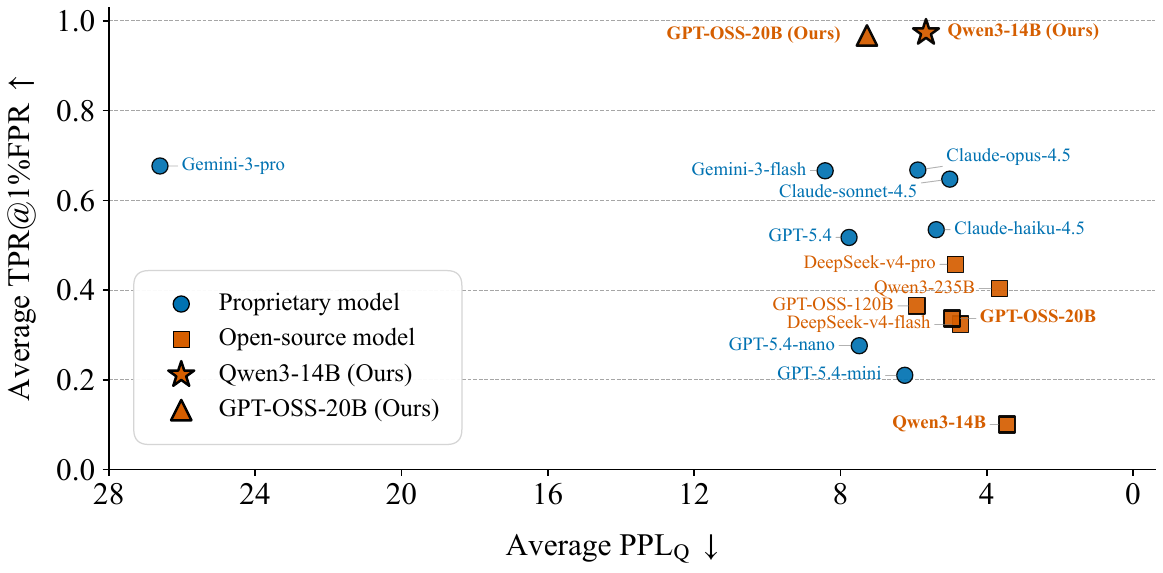}
\captionsetup{width=0.9\linewidth}
\vspace{-0.5em}
\caption{TPR@$1\%$FPR versus perplexity of 14 evaluated frontier proprietary and open-source LLMs on \icwb. Each point represents a model's average performance across three ICW instruction families. Our method raises the average TPR@$1\%$FPR of Qwen3-14B from $0.100$ to $0.974$ and of GPT-OSS-20B from $0.337$ to $0.968$ while maintaining low perplexity, achieving a more favorable trade-off than all the frontier proprietary LLMs we evaluate.}
\label{fig:teaser}
\end{figure}

\end{abstract}

\section{Introduction}
The instruction-following (IF) capability of large language models (LLMs) has been studied primarily as a lever for controllability \cite{zhou2023instruction,jiang2024followbench,zheng2023judging}, powering applications such as formatted generation \cite{willard2023efficient,zheng2024sglang} and tool-using agents \cite{yao2022react,schick2023toolformer,patil2024gorilla,liu2026convexbench}. We study a different use of this capability: \emph{watermarking via instruction}. Concretely, an \emph{in-context watermark} (ICW) instruction~\cite{liu2025context} asks the model to produce an answer that carries a statistically detectable signal.

Watermarking is an important and reliable tool for detecting AI misuse. Most existing LLM watermarking techniques are designed for model owners and require access to the internal sampling process, which is not available to the general public \cite{kirchenbauer2023watermark,zhao2023provable,liu2024adaptive,he2024theoretically,dathathri2024scalable,gumbel2023,christ2023undetectable, liu2024survey, liu2023semantic, liu2023unforgeable}. However, most LLM providers do not publicly disclose the use of watermarks due to different concerns \cite{liu2025position}. This makes AI misuse detection difficult for the general public or third-party stakeholders. To address this problem, \citet{liu2025context} proposes the concept of ICWs and studies their application to detecting AI-generated reviews for academic conferences by injecting the ICW instruction into the manuscript submissions. Moreover, ICML 2026 adopts a similar ICW-based detection for identifying AI-generated reviews \cite{icml2026icw,rao2025detecting}, where 795 reviews ($\sim1\%$ of all reviews) are detected as AI-generated.

The reliability of ICW rests on the LLM's ability to follow the embedded instruction, yet how well current LLMs do so, and at what cost to answer quality, has not been measured. We address this gap with \icwb, a benchmark of three verifiable ICW instruction families (Table~\ref{tab:icw-instructions}): Token-Set Preference (TSP), Word-Initial Preference (WIP), and Sentence Acrostic (SA). Each family is designed to be \emph{general across queries}, \emph{automatically verifiable}, \emph{quality-preserving}, and \emph{robust to editing}. Unlike prior IF benchmarks \cite{zhou2023instruction} that score obedience alone, \icwb~scores both whether the response follows the instruction (carries the watermark signal) and whether it remains a high-quality answer to the query. We evaluate 14 frontier proprietary and open-source LLMs on \icwb, including GPT-5.4 \cite{openai2026gpt54}, Gemini-3-pro \cite{google2025gemini3pro}, Claude-opus-4.5 \cite{anthropic2025claudeopus45}, and DeepSeek-v4-pro \cite{deepseek2026v4}. We observe two contrasting failure modes: \emph{under-following}, where no evaluated model exceeds $0.66$ AUC on TSP, and \emph{over-following}, where strong watermark detectability comes at a substantial cost to output quality. For example, Gemini-3-pro reaches $1.000$ AUC on WIP, while perplexity rises from $4.33$ to $67.61$.




Improving ICW IF capability presents a bootstrapping challenge: the models that need training are precisely those that cannot reliably produce instruction-compliant responses in the first place. Although the automatic verifier makes reinforcement learning with verifiable rewards (RLVR) a natural training framework, effective RLVR requires an initial policy that can already generate sufficiently high-reward rollouts, which weak ICW IF models may fail to provide. Obtaining such a cold start is itself non-trivial. Direct sampling from the base model yields few instruction-compliant responses, manual annotation is costly and difficult to scale, and distillation from stronger LLMs is unreliable because even frontier models exhibit limited ICW IF ability. Moreover, ICW compliance is inherently distributional: the watermark signal is encoded through aggregate token statistics rather than a particular target sequence. Consequently, even given compliant responses, standard supervised fine-tuning (SFT) discards the distribution-level structure that carries the ICW signal. This raises a central question:

\begin{center}
    \it How can we bootstrap reliable ICW IF capability in an LLM with weak initial ICW IF ability, without relying on stronger teacher models or human supervision?
\end{center}

In this paper, we propose a general \emph{self-contained} method that runs on the base LLM alone: no distillation from a stronger model, no manual annotation, no strong pre-existing ICW IF ability required. Our intuition is to convert ICW IF into an instruction-equivalent decoding-time logit perturbation on the base LLM. It resolves both cold-start training problems: (1) it provides a method to synthesize instruction-compliant data from the base LLM itself, and (2) it provides a distribution-level supervision signal that is not available in SFT. Specifically, we propose a two-stage training method. Stage one, \emph{self-distillation with logits perturbation} (SDLP), pairs a frozen teacher with a trainable student, both copies of the base LLM. The teacher receives only the query and produces an instruction-aligned distribution via logit perturbation; the student receives both the ICW instruction and the query without any perturbation, and is trained to match the teacher's distribution. Stage two applies GRPO \cite{shao2024deepseekmath} with the automated verifier as the reward. Applied to two open-source LLMs, Qwen3-14B and GPT-OSS-20B, our method raises their average TPR@1\%FPR across the three ICW instruction families from $0.100$ to $0.974$ and from $0.337$ to $0.968$, respectively, and achieves a more favorable IF-quality trade-off than the frontier LLMs we evaluate.

We summarize our contributions as follows:
\begin{itemize}[leftmargin=1.2em, itemsep=2pt, topsep=2pt]
\item We evaluate 14 frontier proprietary and open-source LLMs on \icwb~and identify the trade-off of frontier LLMs between following ICW instructions and preserving generation quality.
\item We propose a self-contained two-stage training method to improve LLMs' ability to follow ICW instructions while maintaining generation quality, without external supervision (stronger teacher model, human annotation, pre-existing capability).
\item We validate our method on two open-source LLMs with weak initial ICW IF ability, Qwen3-14B and GPT-OSS-20B. Our trained models achieve average TPR@$1\%$FPR of $0.974$ and $0.968$, respectively, across the three instruction families while maintaining high response quality, resulting in a more favorable ICW IF-quality trade-off than the frontier LLMs we evaluate.
\end{itemize}

\section{Related Works}
\textbf{LLM Watermarks.} AI-generated content detection is crucial for mitigating risks of AI misuse \cite{an2026reinforcement,liu2025dataset,an2025defending,liu2024image,yang2025tokenpure,yang2023survey,zhao2024sokwatermark,chang2024postmark,zhu2024duwak,zhang2024remark,hou2023semstamp,zhao2023protecting,ci2024ringid,huang2025rlcracker,he2026fundamental}. Post-hoc detectors \cite{adam2026gptzero,mitchell2023detectgpt,hu2023radar} train classifiers to identify patterns in generated text, but are vulnerable to high false positive rates \cite{sadasivan2023can, liang2023gpt}, which limits their reliability. LLM watermarking methods actively embed a verifiable signal into AI-generated text, usually providing high true positive rates and controllable false positive rates \cite{zhao2023provable,he2024theoretically,zhao2024permute,he2025distributional,li2025statistical,li2024robust,giboulot2024watermax,fernandez2023three,qu2024provably,chen2025improved,cui2026mc,chen2026more,wu2025analyzing}. Most existing generative AI watermarking techniques are designed for model owners and require access to the internal sampling process, which is not available to the general public \cite{kirchenbauer2023watermark,liu2024adaptive,dathathri2024scalable,gumbel2023,christ2023undetectable,kuditipudi2023robust}. However, the application of watermarks depends on the LLM providers, which significantly limits the real-world adoption of LLM watermarking. A natural alternative is to shift control of watermarking from the model owner to the user or third party: the LLM itself becomes a watermarking tool that requires no access to its internals. ICWs \cite{liu2025context,rao2025detecting} realize this idea by only using the ICW instructions. Specifically, \citet{liu2025context} proposes a general ICW framework and investigates four ICW strategies that watermark LLM-generated text at different linguistic granularities, which are broadly applicable across generation tasks. \citet{rao2025detecting} targets only the peer-review scenario and instructs the LLMs to insert stochastically chosen trigger phrases into the generated reviews whose presence serves as the detection signal. In this paper, we focus on general ICW methods, where instructions apply across arbitrary queries and can be evaluated as a general LLM capability. We benchmark frontier LLMs on a suite of ICW instructions, characterize where they fail to comply, and propose a training method to close this gap.

\textbf{LLM Instruction-Following Capabilities.} The IF ability of LLMs has been extensively studied and evaluated \cite{zhou2023instruction, mu2023can, jiang2024followbench, qin2024infobench}, and improved through instruction tuning \cite{longpre2023flan,sanh2021multitask,wang2023self}, RLHF \cite{ouyang2022training}, and preference optimization \cite{rafailov2023direct}. ICW instructions, however, differ from the constraints studied in this line of work along two axes. First, prior benchmarks score \emph{per-response} compliance with surface-level constraints, e.g., length, format, keyword inclusion, or style \cite{zhou2023instruction,jiang2024followbench}; in contrast, an ICW instruction imposes a constraint on the \emph{aggregate statistics} of a response (e.g., token frequencies), whose satisfaction is verified by a statistical test rather than by a deterministic surface check. As a result, a model can satisfy the instruction in a distributional sense, by slightly biasing token-level choices throughout the response, even though no individual token is required to take a specific value. Second, prior IF benchmarks score compliance largely in isolation from answer quality, whereas ICW instructions require \emph{both} detectability and quality on the same query; \icwb~therefore scores both, exposing an IF-quality trade-off that single-axis evaluation does not surface. Consequently, existing IF benchmarks do not evaluate ICW instruction-following ability, and standard IF training pipelines~\cite{dong2024self} do not directly transfer.

\section{ICWBench}
\label{sec:icwbench}

To evaluate whether an LLM can comprehend and execute ICW instructions, inspired by~\cite{liu2025context}, we introduce \icwb, a compact benchmark built around three families of \textit{verifiable} ICW instructions. Each family is parameterized so that a template yields watermarked text with different keys, and each is paired with a statistical test for verification. In the rest of this section, we describe the instruction design and evaluation metrics in detail.

\subsection{ICW Instruction}
\label{sec:icw-instruction}

\paragraph{Problem formulation.}
Let $\mathcal{M}$ denote an LLM and $q$ a user query. An ICW instruction is a pair $(\psi, \kappa)$, where $\psi$ is an instruction family, $\pi \in \Pi_\psi$ is an instruction template in that family, and $\kappa \in \mathcal{K}_\psi$ is a secret key. Prepending the instantiated instruction to the query yields the response $y \leftarrow \mathcal{M}(\pi(\kappa), q)$. Each family is paired with a verifier $V_\psi : \mathcal{Y} \times \mathcal{K}_\psi \to \mathbb{R}$ that maps a response to a test statistic. The benchmark asks whether $V_\psi(y; \kappa)$ exceeds a detection threshold while $y$ remains an answer to $q$.

\paragraph{Design principles.}
We select instructions according to four properties. (1)~\emph{Generalizability}: the template should apply to arbitrary queries rather than a narrow task family. (2)~\emph{Verifiability}: the induced signal in $y$ should be detectable by an automatic verifier. (3)~\emph{Quality preservation}: complying with the instruction should leave the response a faithful answer to $q$. (4)~\emph{Robustness}: the generated text should be robust to editing operations.

\paragraph{Instruction families.}
These principles lead us to focus on ICW instructions that are content-agnostic and can be applied across query domains. This differs from the phrase-insertion approach used for LLM-generated review detection by \citet{rao2025detecting}, which instructs the model to insert preselected trigger phrases into a review. Such an approach is well suited to its application-specific detection setting, but largely reduces IF to the inclusion of a small number of strings. Our goal is instead to evaluate ICW IF as a general LLM capability, including instructions that constrain statistical or structural properties of the response while leaving its semantic content unconstrained.
\icwb~instantiates three families, summarized in Table~\ref{tab:icw-instructions}. The three families impose constraints at progressively coarser granularities: Token-Set Preference (TSP) biases which tokens may appear, Word-Initial Preference (WIP) biases the initial letter of each word, and Sentence Acrostic (SA) requires sentence-initial letters to spell a target string in order. All three are independent of the query domain and admit automatic statistical verification. As shown in~\cite{liu2025context}, they cover distinct forms of ICW IF while satisfying our criteria of generalizability, verifiability, quality preservation, and robustness.

\begin{table}[t]
\centering
\caption{The three families of ICW instructions in \icwb, with a brief description. Each family is parameterized by $\kappa$, yielding distinct instances per family.}
\label{tab:icw-instructions}
\scriptsize
\begin{tabularx}{\textwidth}{p{0.2\linewidth} l X}
\toprule
\textbf{Instruction Type} & \textbf{Parameters $\kappa$} & \textbf{Description} \\
\midrule
Token-Set Preference\newline(TSP) & $T = \{t_1, t_2, \dots, t_k\}$ & In answering a query naturally, increase the use of tokens from the given token set $T$ in the generated text without compromising quality. \\
\addlinespace
Word-Initial Preference\newline(WIP) & $L = \{\ell_1, \ell_2, \dots, \ell_k\}$ & In answering a query naturally, increase the use of words starting with the letters in $L$ in the generated text without compromising quality. \\
\addlinespace
Sentence Acrostic\newline(SA) & $S = s_1 s_2 \cdots s_k$ & In answering a query naturally, structure sentences so their initial letters sequentially spell the given string $S$ without compromising quality. \\
\bottomrule
\end{tabularx}
\vspace{-1em}
\end{table}

\subsection{Evaluation Metrics}
\label{sec:eval-metrics}

For a family $\psi$ with parameter $\kappa$, let $n$ be the number of scored units in the response $y$ (tokens for TSP and word-initials for WIP), and let $X$ count how many of these positions are instruction-aligned. Under the null hypothesis $H_0$ that $y$ is independent of ICW instructions, let $p_0$ denote the probability that a scored unit is instruction-aligned. The verification score is the $z$-statistic
\begin{equation}
\label{eq:zstat}
z \;=\; \frac{X - n\, p_0}{\sqrt{n\, p_0\,(1 - p_0)}}.
\end{equation}
A larger $z$ provides stronger evidence against $H_0$ and thus that LLMs follow the instruction, and $y$ carries the watermark signal. TSP and WIP admit this closed form in \eqref{eq:zstat}. SA instead scores a longest-common-subsequence match against a permutation null, which generalizes \eqref{eq:zstat}. We instantiate the details of $(X, n, p_0)$ per family in Appendix~\ref{appendix:perfamily}.

\section{Learning to Follow ICW Instructions}

The automatic verifier~\eqref{eq:zstat} of ICW instructions enables a natural reward signal for training LLMs to follow ICW instructions through reinforcement learning. However, effective RLVR typically requires an initial policy that can sample rollouts with high reward, a condition that base LLMs may not satisfy on ICW instructions (see Figure~\ref{fig:teaser}). Therefore, an effective cold-start training strategy is needed to give RLVR a better starting point. The cold-start training faces two challenges: (i) how to automatically synthesize high-quality training data; (ii) how to design a training method to effectively use the synthesized data.

\begin{figure}[t]
\centering
\includegraphics[width=1\linewidth]{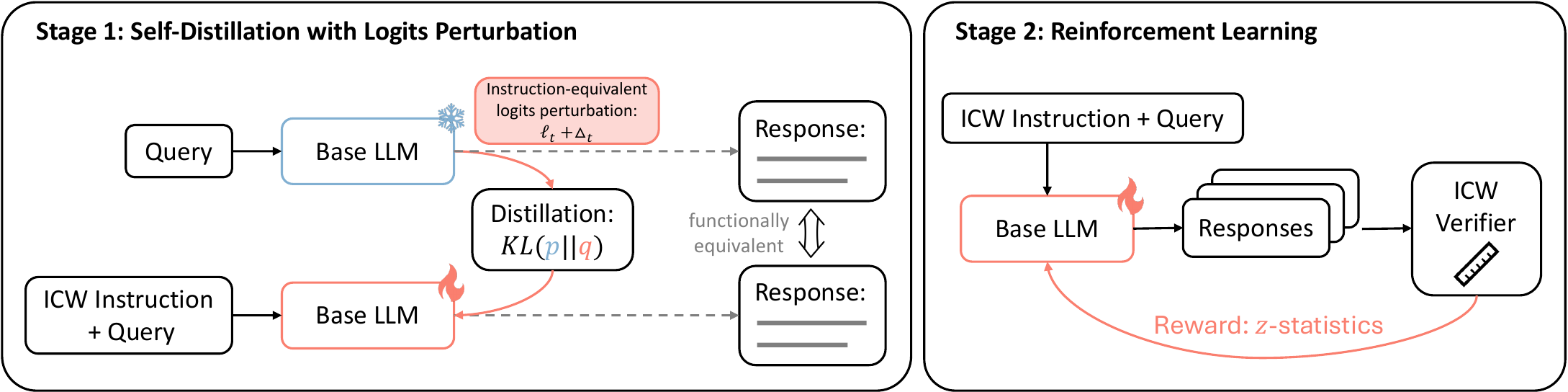}
\captionsetup{width=1\linewidth}
\caption{Our two-stage training workflow. Stage 1 (self-distillation with logits perturbation): A frozen base LLM, given only the query and with an ICW instruction-equivalent logits perturbation at decoding time, serves as the teacher; the student model (initially the same as the teacher model), conditioned on the ICW instruction and query, is trained to match the teacher's output distribution. Stage 2 (RL): The Stage-1 model is then refined by the reward of the automatic ICW verifier $V_\psi(\,\cdot\,;\kappa)$.}
\label{fig:method_workflow}
\vspace{-0.5em}
\end{figure}

To address these problems, we propose a two-stage approach (Figure~\ref{fig:method_workflow}). In the first stage, we propose \emph{self-distillation with logits perturbation} (SDLP), which bootstraps an initial policy that already produces aligned, high-quality rollouts, supplying RLVR with a better starting point. The second stage applies Group Relative Policy Optimization (GRPO) \cite{shao2024deepseekmath} on top of the SDLP-initialized policy, using \eqref{eq:zstat} as the reward to further improve IF ability. We describe SDLP in Section~\ref{sec:sdlp} and RLVR in Section~\ref{sec:rlvr}.

\subsection{Training Data Synthesis}
\label{sec:tds}

A high-quality ICW IF dataset is crucial for cold-start training, yet automatically synthesizing such data at scale is non-trivial. Direct sampling from the base LLM yields few instruction-aligned responses, since base LLMs do not reliably follow all ICW instructions. Distillation from a stronger LLM is also unreliable: frontier proprietary models also show limited ability to follow ICW instructions. The way forward comes from a property of ICW verification: it depends on the response's aggregate token statistics, not on a specific surface form.

The core idea is to enforce the IF at decoding time via instruction-equivalent logit perturbation, rather than rely on the base LLM to follow the ICW instruction.
Concretely, given a query $q$, the base LLM $\mathcal{M}$ generates a response $y$ autoregressively, one token $y_t$ at a time. At each decoding step $t$, let $\boldsymbol{\ell}_t \in \mathbb{R}^{|\mathcal{V}|}$ denote the logits produced by $\mathcal{M}$ over vocabulary $\mathcal{V}$, conditioned on the prefix $q$ and $y_{<t}$. We add an instruction-dependent perturbation $\boldsymbol{\Delta}_t(\psi; \kappa) \in \mathbb{R}^{|\mathcal{V}|}$ to $\boldsymbol{\ell}_t$ and sample the next token from the perturbed distribution:
\begin{equation}
\label{eq:logit-perturb}
y_t \;\sim\; \mathrm{softmax}\!\left(\boldsymbol{\ell}_t + \boldsymbol{\Delta}_t(\psi; \kappa)\right).
\end{equation}
The perturbation $\boldsymbol{\Delta}_t$ is chosen to up-weight tokens that move $y$ toward a higher verifier score $V_\psi(y; \kappa)$. For example, under the TSP family with parameter $T \subset \mathcal{V}$, a natural choice is
\begin{equation*}
\boldsymbol{\Delta}_t[v] \;=\; \delta \cdot \mathbf{1}[v \in T], \qquad \delta > 0,
\end{equation*}
which uniformly boosts the logits of tokens in $T$ and thereby raises the expected count of $T$-tokens in $y$, the very quantity scored by the $z$-statistic in~\eqref{eq:zstat}. The concrete design of $\boldsymbol{\Delta}_t$ should be adapted to the specific ICW instruction family. For details of the perturbation design for three ICW instruction families, see Appendix~\ref{appendix:perfamily}.

Repeating this procedure across different queries $q$ and instruction parameters $(\psi, \kappa)$ yields a synthetic cold-start dataset
\begin{equation}
\label{eq:cs-dataset}
\mathcal{D}_{\text{cs}} \;=\; \mathcal{D}_{H_1} \cup \mathcal{D}_{H_0} \;=\; \big\{\, \big(\psi,\, \kappa,\, q^{(i)},\, y^{(i)}\big) \,\big\}_{i=1}^{N} \cup \big\{\, \big(\hat{q}^{(i)},\, \hat{y}^{(i)}\big) \,\big\}_{i=1}^{M}.
\end{equation}
The watermarked part $\mathcal{D}_{H_1}$ collects the perturbed samples: by construction, every $y^{(i)}$ follows the ICW instruction for $(\psi, \kappa)$ scored by $V_{\psi}(\,\cdot\,;\kappa)$, where $\mathcal{M}$ does not see the instruction in its context.  The unwatermarked part $\mathcal{D}_{H_0}$ collects pairs $(\hat{q}^{(i)}, \hat{y}^{(i)})$, where $\hat{y}^{(i)}$ is text generated under null hypothesis. The evaluation of $\mathcal{D}_{\text{cs}}$ can be found in Appendix~\ref{appendix:dcs}, showing that it achieves high reward while maintaining good quality.

\subsection{Self-Distillation with Logits Perturbation}
\label{sec:sdlp}

A natural way to use $\mathcal{D}_{\text{cs}}$ is SFT on the realized sequences $y^{(i)}$. However, the ICW signal is encoded in the \emph{distribution} of the response, spread across many token choices rather than any single emitted token. Standard SFT discards this distribution-level information at each decoding step.

We therefore distill the ICW distribution, rather than only the realized token. The remaining question is how to obtain a teacher distribution that already follows the ICW instruction. The logit perturbation $\boldsymbol{\Delta}_t(\psi; \kappa)$ from Section~\ref{sec:tds} directly defines such a distribution and can therefore serve as the teacher.

\paragraph{Teacher -- student models setup.}
Both teacher and student models are copies of the base LLM $\mathcal{M}$. The teacher model remains frozen; the student model, denoted $\mathcal{M}_\theta$, is updated during training. Moreover, the two models receive different inputs at each decoding step $t$:
\begin{align*}
\boldsymbol{\ell}_t^{\mathcal{T}} \;\leftarrow\; \mathcal{M}\!\left(\,\cdot \,\big|\, q,\, y_{<t}\right), ~~~ \boldsymbol{\ell}_t^{\mathcal{S}} \;\leftarrow\; \mathcal{M}_\theta\!\left(\,\cdot \,\big|\, \pi(\kappa),\, q,\, y_{<t}\right).
\end{align*}
Our core idea is to distill the teacher distribution into the student, so that given the ICW instruction $\pi(\kappa)$, the student learns to reproduce the distribution induced by the following perturbed decoding.

\paragraph{Self-distillation loss.}
The teacher distribution at step $t$ is
\begin{equation}
\label{eq:teacher-dist}
p_t^{\mathcal{T}} \;=\; \mathrm{softmax}\!\left(\boldsymbol{\ell}_t^{\mathcal{T}} + \boldsymbol{\Delta}_t\right),
\qquad
\boldsymbol{\Delta}_t \;=\;
\begin{cases}
\boldsymbol{\Delta}_t(\psi; \kappa) & \text{for samples in } \mathcal{D}_{H_1}, \\
\boldsymbol{0} & \text{for samples in } \mathcal{D}_{H_0}.
\end{cases}
\end{equation}
On $\mathcal{D}_{H_0}$, the target is the base LLM's natural output distribution (no instruction signal); on $\mathcal{D}_{H_1}$, it is the perturbed distribution. The student distribution is $p_t^{\mathcal{S}} = \mathrm{softmax}(\boldsymbol{\ell}_t^{\mathcal{S}})$. The SDLP loss is:
\begin{equation}
\label{eq:sdlp-loss}
\mathcal{L}_{\mathrm{SDLP}}(\theta)
\;=\; \mathbb{E}_{(q, y) \sim \mathcal{D}_{\text{cs}}}\!\left[\, \sum_{t=1}^{|y|}
\mathrm{KL}\!\left(\, \tilde{p}_t^{\mathcal{T}} \,\big\|\, \tilde{p}_t^{\mathcal{S}} \,\right) \right],
\end{equation}
where $\mathrm{KL}\!\left(\tilde{p}_t^{\mathcal{T}}\big\|\tilde{p}_t^{\mathcal{S}}\right)
= \sum_{i} \tilde{p}_t^{\mathcal{T}}(i) \log \frac{\tilde{p}_t^{\mathcal{T}}(i)}{\tilde{p}_t^{\mathcal{S}}(i)}$ is the KL divergence, and $\tilde{p}_t^{\mathcal{T}}, \tilde{p}_t^{\mathcal{S}}$ are top-$k$ truncated distributions defined below.

\paragraph{Top-$k$ truncation.} In practice, low-probability tail of $p_t^{\mathcal{T}}$ contributes little meaningful training signal. Therefore, we restrict the KL term in \eqref{eq:sdlp-loss} to the top-$k$ tokens of the teacher distribution. Let $\mathcal{I}_t \;=\; \mathrm{TopK}\!\left(p_t^{\mathcal{T}},\, k\right) \;\subset\; \mathcal{V}$ denote the $k$ token indices with the largest probability under $p_t^{\mathcal{T}}$. We restrict \emph{both} $p_t^{\mathcal{T}}$ and $p_t^{\mathcal{S}}$ to this same support $\mathcal{I}_t$ and renormalize:
\begin{equation}
\label{eq:trunc-dists}
\tilde{p}_t^{\mathcal{T}}[v] \;=\; \frac{p_t^{\mathcal{T}}[v] \cdot \mathbf{1}[v \in \mathcal{I}_t]}{\sum_{v' \in \mathcal{I}_t} p_t^{\mathcal{T}}[v']},
\qquad
\tilde{p}_t^{\mathcal{S}}[v] \;=\; \frac{p_t^{\mathcal{S}}[v] \cdot \mathbf{1}[v \in \mathcal{I}_t]}{\sum_{v' \in \mathcal{I}_t} p_t^{\mathcal{S}}[v']}.
\end{equation}

\subsection{Reinforcement Learning with Verifiable Rewards}
\label{sec:rlvr}

After SDLP, the policy $\mathcal{M}_\theta$ learns to produce rollouts that follow ICW instructions and earn high rewards. We further improve $\mathcal{M}_\theta$ with GRPO~\cite{shao2024deepseekmath}, using the verifier $V_\psi(\,\cdot\,;\kappa)$ from Section~\ref{sec:eval-metrics} as the reward and rolling out only on $\mathcal{D}_{H_1}$. For each $(\psi, \kappa, q) \sim \mathcal{D}_{H_1}$, we sample $G$ rollouts $\{y^{(i)}\}_{i=1}^{G}$ from the current policy $\mathcal{M}_{\theta_{\mathrm{old}}}$, score them by $r_i = V_\psi(y^{(i)}; \kappa)$, and compute group-relative advantages
\begin{equation}
\label{eq:grpo-advantage}
A_i \;=\; \frac{r_i - \bar{r}}{\sigma_r},
\qquad
\bar{r} \;=\; \frac{1}{G}\sum_{j=1}^{G} r_j,
\qquad
\sigma_r \;=\; \sqrt{\frac{1}{G}\sum_{j=1}^{G}(r_j - \bar{r})^2 + \varepsilon}.
\end{equation}
Let $\rho_t^{(i)}(\theta) = \mathcal{M}_\theta\!\big(y_t^{(i)} \,\big|\, \pi(\kappa),\, q,\, y_{<t}^{(i)}\big) \,\big/\, \mathcal{M}_{\theta_{\mathrm{old}}}\!\big(y_t^{(i)} \,\big|\, \pi(\kappa),\, q,\, y_{<t}^{(i)}\big)$ denote the per-token importance ratio. We maximize:
\begin{equation}
\label{eq:rlvr-loss}
\mathcal{J}_{\mathrm{RLVR}}(\theta) = \mathbb{E}_{(\psi,\kappa,q) \sim \mathcal{D}_{H_1}}\!\left[\frac{1}{G}\sum_{i=1}^{G} \frac{1}{|y^{(i)}|}\sum_{t=1}^{|y^{(i)}|} \min\!\Big(\rho_t^{(i)}(\theta)A_i,\; \mathrm{clip}\big(\rho_t^{(i)}(\theta),\, 1-\epsilon,\, 1+\epsilon\big) A_i \Big) \right],
\end{equation}
where $\epsilon$ is the clipping range.

\section{Experiments}

\subsection{Experiment Settings}

\paragraph{Models \& Datasets \& Baselines.}
\textit{Evaluation queries.} We sample $500$ queries from LFQA\footnote{\url{https://huggingface.co/datasets/vblagoje/lfqa}}, a long-form question-answering dataset whose open-ended queries admit substantial response variation and thus expose the IF versus quality trade-off more clearly than short factual questions. The same $500$ queries are reused across all three instruction families and all evaluated models.

\textit{Training data.} The training set is built from LFQA queries disjoint from the $500$ evaluation queries. For positives ($\mathcal{D}_{H_1}$), we use $3379$, $1865$, $2906$ queries for TSP, WIP, and SA, respectively, and synthesize one instruction-aligned response per query using the data synthesis method in Section~\ref{sec:tds}. For negatives ($\mathcal{D}_{H_0}$), we sample $1000$, $800$, $800$ additional queries per family and generate a single response per query from the base model with the query alone as input. Together, the positive and negative examples form the cold-start dataset $\mathcal{D}_{\mathrm{cs}}$ in~\eqref{eq:cs-dataset}.

\textit{Baselines \& Models.} We benchmark $14$ frontier LLMs spanning two groups. \emph{Proprietary}: GPT-5.4 \cite{openai2026gpt54}, GPT-5.4-mini \cite{openai2026gpt54mini}, GPT-5.4-nano \cite{openai2026gpt54nano}, Gemini-3-pro \cite{google2025gemini3pro}, Gemini-3-flash \cite{google2025gemini3flash}, Claude-opus-4.5 \cite{anthropic2025claudeopus45}, Claude-sonnet-4.5 \cite{anthropic2025claudesennet45}, Claude-haiku-4.5 \cite{anthropic2025claudehaiku45}. \emph{Open-source}: DeepSeek-v4-pro, DeepSeek-v4-flash \cite{deepseek2026v4}, Qwen3-235B, Qwen3-14B \cite{qwen3technicalreport}, GPT-OSS-120B, GPT-OSS-20B \cite{openai2025gptoss}. The same $14$ models serve as the baselines against which we compare our trained models; alternative training methods (SFT, direct RL) are reported in the ablation study (Table~\ref{tab:method_ablation}). All models are queried with default decoding parameters and a generation cap of $600$ new tokens.

\paragraph{Implementation Details.} We train Qwen3-14B and GPT-OSS-20B, two open-source models with weak initial ICW IF ability among all LLMs we evaluate. All experiments are conducted on NVIDIA B200 GPUs and RTX PRO 6000 Blackwell GPUs. Full implementation details can be found in Appendix~\ref{appendix:impl}.

\paragraph{Evaluation Metrics.}
We evaluate each model along two axes: ICW IF performance and response quality. All metrics are reported per instruction family.

\textit{ICW IF Performance.}
For each query $q$, we collect a positive response $y^+$ generated under $\pi(\kappa)$ and $q$ and a negative response $y^-$ generated under $q$ alone, and score both with the $z$-statistic in~\eqref{eq:zstat}. We report ROC-AUC and TPR@$1\%$FPR; the latter measures the true positive rate at a controlled $1\%$ false positive rate. 

\textit{Response Quality.}
We assess response quality using both perplexity (PPL) and LLM-as-a-Judge. PPL is computed on $y^+$ using two external reference models, Nemotron-3-Super-120B \cite{nvidia_nemotron_3_2025} and Qwen3-235B, conditioned on the query $q$ as the prefix; lower PPL indicates that the watermarked response remains a likely continuation of the query under these reference models. For LLM-as-a-Judge, we provide \texttt{gpt-4o-mini} with the query and generated response and ask it to assign an integer score from $1$ to $5$ along four dimensions: Relevance, Clarity, Informativeness, and Factual Correctness.

\begin{table}[t]
\centering
\caption{ICW instruction-following on \icwb: ROC-AUC and TPR@$1\%$FPR on each instruction family, for 14 frontier LLMs that were not trained for ICW IF and our two trained models. Averaged across the three families, our method raises TPR@$1\%$FPR from $0.100$ to $0.974$ on Qwen3-14B and from $0.337$ to $0.968$ on GPT-OSS-20B. Answer quality for the same responses is reported separately in Table~\ref{tab:text_quality}. $^{\dagger}$~marks the two backbones we train.}
\vspace{0.5em}
\label{tab:main_results}
\scriptsize
\setlength{\tabcolsep}{7.23pt}
\begin{tabular}{llcccccc}
\toprule
& \multirow{2}{*}{\textbf{Models}} & \multicolumn{2}{c}{\textbf{TSP}} & \multicolumn{2}{c}{\textbf{WIP}} & \multicolumn{2}{c}{\textbf{SA}} \\
\cmidrule(lr){3-4} \cmidrule(lr){5-6} \cmidrule(lr){7-8}
& & AUC $\uparrow$ & TPR@1\%FPR $\uparrow$ & AUC & TPR@1\%FPR & AUC & TPR@1\%FPR \\
\midrule
\multirow{12}{*}{Proprietary}
& Gemini-3-pro        & $0.658$ & $0.030$ & $1.000$ & $1.000$ & $1.000$ & $1.000$ \\ \addlinespace
& Gemini-3-flash      & $0.646$ & $0.040$ & $0.998$ & $0.958$ & $1.000$ & $1.000$ \\ \addlinespace
& Claude-opus-4.5     & $0.597$ & $0.022$ & $0.998$ & $0.983$ & $1.000$ & $0.998$ \\ \addlinespace
& Claude-sonnet-4.5   & $0.487$ & $0.011$ & $0.996$ & $0.948$ & $1.000$ & $0.983$ \\ \addlinespace
& Claude-haiku-4.5    & $0.600$ & $0.098$ & $0.961$ & $0.642$ & $0.989$ & $0.864$ \\ \addlinespace
& GPT-5.4             & $0.538$ & $0.020$ & $0.933$ & $0.549$ & $0.998$ & $0.983$ \\ \addlinespace
& GPT-5.4-mini        & $0.527$ & $0.020$ & $0.692$ & $0.059$ & $0.913$ & $0.551$ \\ \addlinespace
& GPT-5.4-nano        & $0.462$ & $0.000$ & $0.529$ & $0.004$ & $0.973$ & $0.824$ \\
\midrule
& DeepSeek-v4-pro     & $0.579$ & $0.120$ & $0.841$ & $0.260$ & $0.999$ & $0.992$ \\ \addlinespace
& DeepSeek-v4-flash   & $0.576$ & $0.030$ & $0.832$ & $0.277$ & $0.957$ & $0.665$ \\ \addlinespace
& Qwen3-235B & $0.591$ & $0.020$ & $0.837$ & $0.252$ & $0.995$ & $0.937$ \\ \addlinespace
\rowcolor{gray!12}
\cellcolor{white} & Qwen3-14B$^{\dagger}$           & $0.564$ & $0.020$ & $0.583$ & $0.036$ & $0.830$ & $0.245$ \\ \addlinespace
& GPT-OSS-120B        & $0.625$ & $0.020$ & $0.781$ & $0.166$ & $0.983$ & $0.908$ \\ \addlinespace
\rowcolor{gray!12}
\cellcolor{white}\multirow{-9}{*}{Open-source} & GPT-OSS-20B$^{\dagger}$         & $0.575$ & $0.010$ & $0.568$ & $0.015$ & $0.998$ & $0.985$ \\
\midrule
\rowcolor{blue!8}
\cellcolor{white} & \textbf{Qwen3-14B} & $\mathbf{0.993}$ & $\mathbf{0.941}$ & $\mathbf{0.999}$ & $\mathbf{0.996}$ & $\mathbf{0.999}$ & $\mathbf{0.985}$ \\ \addlinespace
\rowcolor{blue!8}
\cellcolor{white}\multirow{-2}{*}{Ours} & \textbf{GPT-OSS-20B} & $\mathbf{0.983}$ & $\mathbf{0.922}$ & $\mathbf{0.992}$ & $\mathbf{0.989}$ & $\mathbf{0.999}$ & $\mathbf{0.992}$ \\
\bottomrule
\end{tabular}
\end{table}

\subsection{Main Results}

\subsubsection{Evaluation of frontier models on \icwb}

We evaluate the $14$ frontier LLMs on \icwb, reporting ICW IF performance in Table~\ref{tab:main_results} and response quality in Table~\ref{tab:text_quality}.

\paragraph{ICW IF Performance.}
IF performance varies sharply across both instruction families and models. SA is the easiest family: every evaluated LLM reaches AUC $\geq 0.83$, and five proprietary models exceed TPR@$1\%$FPR $= 0.98$. WIP is harder but tractable for some proprietary models: Gemini-3-pro, Gemini-3-flash, Claude-opus-4.5, and Claude-sonnet-4.5 all exceed TPR@$1\%$FPR $= 0.94$, while other models stay below TPR@$1\%$FPR $= 0.65$ (e.g., GPT-5.4 at $0.549$) and $8$ of them below $0.30$ (e.g., DeepSeek-v4-pro at $0.260$, GPT-OSS-120B at $0.166$). TSP defeats every frontier model: the strongest is DeepSeek-v4-pro at TPR@$1\%$FPR $= 0.120$, and the strongest TSP AUC across all evaluated LLMs is $0.658$.

\begin{table}[t]
\centering
\caption{Response quality on \icwb, measured on the same responses as Table~\ref{tab:main_results}. PPL$_{\mathrm{Q}}$ and PPL$_{\mathrm{N}}$ are perplexities computed by Qwen3-235B and Nemotron-3-Super-120B, two reference models from different families. R, C, I, and FC are LLM-as-a-Judge scores (\texttt{gpt-4o-mini}) for Relevance, Clarity, Informativeness, and Factual Correctness, each an integer in $1$--$5$; Avg is their mean. Our trained models remain comparable to their backbones and to the frontier baselines on both metrics. $^{\dagger}$~marks the two backbones we train.}
\vspace{0.5em}
\label{tab:text_quality}
\scriptsize
\setlength{\tabcolsep}{1.2pt}
\resizebox{\textwidth}{!}{%
\begin{tabular}{llccccccccccccccccccccc}
\toprule
& \multirow{3}{*}{\textbf{Models}} & \multicolumn{7}{c}{\textbf{TSP}} & \multicolumn{7}{c}{\textbf{WIP}} & \multicolumn{7}{c}{\textbf{SA}} \\
\cmidrule(lr){3-9} \cmidrule(lr){10-16} \cmidrule(lr){17-23}
& & \multicolumn{2}{c}{Perplexity $\downarrow$} & \multicolumn{5}{c}{LLM-as-a-Judge $\uparrow$}
   & \multicolumn{2}{c}{Perplexity $\downarrow$} & \multicolumn{5}{c}{LLM-as-a-Judge $\uparrow$}
   & \multicolumn{2}{c}{Perplexity $\downarrow$} & \multicolumn{5}{c}{LLM-as-a-Judge $\uparrow$} \\
\cmidrule(lr){3-4} \cmidrule(lr){5-9} \cmidrule(lr){10-11} \cmidrule(lr){12-16} \cmidrule(lr){17-18} \cmidrule(lr){19-23}
& & PPL$_{\mathrm{Q}}$ & PPL$_{\mathrm{N}}$ & R & C & I & FC & Avg
   & PPL$_{\mathrm{Q}}$ & PPL$_{\mathrm{N}}$ & R & C & I & FC & Avg
   & PPL$_{\mathrm{Q}}$ & PPL$_{\mathrm{N}}$ & R & C & I & FC & Avg \\
\midrule
\multirow{12}{*}{Proprietary}
& Gemini-3-pro        & $5.55$ & $5.13$ & $4.91$ & $4.59$ & $4.11$ & $4.72$ & $4.58$ & $67.61$ & $53.51$ & $4.40$ & $3.57$ & $3.60$ & $4.18$ & $3.94$ & $6.64$ & $6.38$ & $5.00$ & $4.98$ & $5.00$ & $4.69$ & $4.92$ \\ \addlinespace
& Gemini-3-flash      & $3.35$ & $3.30$ & $5.00$ & $4.86$ & $4.96$ & $4.56$ & $4.85$ & $13.62$ & $11.86$ & $4.98$ & $4.57$ & $4.63$ & $4.47$ & $4.66$ & $8.28$ & $7.81$ & $5.00$ & $4.87$ & $5.00$ & $4.57$ & $4.86$ \\ \addlinespace
& Claude-opus-4.5     & $4.07$ & $3.92$ & $4.92$ & $4.88$ & $4.87$ & $4.72$ & $4.85$ & $7.04$ & $6.34$ & $5.00$ & $4.99$ & $4.98$ & $4.71$ & $4.92$ & $6.54$ & $5.80$ & $5.00$ & $4.97$ & $5.00$ & $4.71$ & $4.92$ \\ \addlinespace
& Claude-sonnet-4.5   & $3.49$ & $3.53$ & $4.89$ & $4.85$ & $4.83$ & $4.64$ & $4.80$ & $6.53$ & $5.88$ & $5.00$ & $4.99$ & $4.99$ & $4.73$ & $4.93$ & $5.01$ & $4.48$ & $5.00$ & $4.98$ & $4.99$ & $4.77$ & $4.93$ \\ \addlinespace
& Claude-haiku-4.5    & $4.47$ & $4.39$ & $4.93$ & $4.83$ & $4.78$ & $4.71$ & $4.81$ & $5.50$ & $5.14$ & $5.00$ & $4.99$ & $4.96$ & $4.77$ & $4.93$ & $6.17$ & $5.64$ & $5.00$ & $4.97$ & $4.98$ & $4.79$ & $4.94$ \\ \addlinespace
& GPT-5.4             & $6.75$ & $6.23$ & $5.00$ & $4.99$ & $4.99$ & $4.93$ & $4.98$ & $8.21$ & $7.54$ & $5.00$ & $4.97$ & $4.98$ & $4.92$ & $4.97$ & $8.33$ & $7.47$ & $5.00$ & $4.94$ & $4.99$ & $4.94$ & $4.97$ \\ \addlinespace
& GPT-5.4-mini        & $5.55$ & $5.16$ & $5.00$ & $4.99$ & $4.97$ & $4.94$ & $4.98$ & $5.94$ & $5.44$ & $4.99$ & $4.99$ & $4.95$ & $4.91$ & $4.96$ & $7.23$ & $6.32$ & $5.00$ & $4.98$ & $4.95$ & $4.92$ & $4.96$ \\ \addlinespace
& GPT-5.4-nano        & $7.21$ & $6.33$ & $4.99$ & $4.98$ & $4.99$ & $4.94$ & $4.98$ & $7.24$ & $6.46$ & $4.98$ & $4.98$ & $4.95$ & $4.96$ & $4.97$ & $8.00$ & $7.01$ & $5.00$ & $4.92$ & $4.99$ & $4.92$ & $4.96$ \\
\midrule
& DeepSeek-v4-pro     & $3.81$ & $3.65$ & $4.97$ & $4.98$ & $4.94$ & $4.76$ & $4.91$ & $4.43$ & $4.36$ & $4.99$ & $4.98$ & $4.89$ & $4.76$ & $4.91$ & $6.34$ & $5.64$ & $5.00$ & $4.97$ & $4.99$ & $4.76$ & $4.93$ \\ \addlinespace
& DeepSeek-v4-flash   & $3.08$ & $3.24$ & $5.00$ & $4.97$ & $4.96$ & $4.71$ & $4.91$ & $4.87$ & $4.74$ & $4.98$ & $4.95$ & $4.75$ & $4.68$ & $4.84$ & $6.22$ & $5.87$ & $4.99$ & $4.89$ & $4.76$ & $4.56$ & $4.80$ \\ \addlinespace
& Qwen3-235B          & $2.78$ & $4.27$ & $5.00$ & $4.96$ & $4.99$ & $4.72$ & $4.92$ & $3.98$ & $5.75$ & $4.97$ & $4.96$ & $4.88$ & $4.82$ & $4.91$ & $4.20$ & $5.84$ & $4.99$ & $4.94$ & $4.98$ & $4.80$ & $4.92$ \\ \addlinespace
\rowcolor{gray!12}
\cellcolor{white} & Qwen3-14B$^{\dagger}$           & $2.80$ & $3.17$ & $4.94$ & $4.95$ & $4.93$ & $4.72$ & $4.89$ & $3.48$ & $3.57$ & $4.96$ & $4.96$ & $4.78$ & $4.72$ & $4.86$ & $4.05$ & $4.21$ & $4.95$ & $4.90$ & $4.74$ & $4.69$ & $4.82$ \\ \addlinespace
& GPT-OSS-120B        & $4.34$ & $3.15$ & $5.00$ & $4.79$ & $4.87$ & $4.65$ & $4.83$ & $6.33$ & $4.64$ & $4.85$ & $4.66$ & $4.66$ & $4.65$ & $4.71$ & $7.06$ & $5.55$ & $4.97$ & $4.57$ & $4.84$ & $4.44$ & $4.71$ \\ \addlinespace
\rowcolor{gray!12}
\cellcolor{white}\multirow{-9}{*}{Open-source} & GPT-OSS-20B$^{\dagger}$         & $4.06$ & $3.19$ & $4.95$ & $4.86$ & $4.91$ & $4.65$ & $4.84$ & $5.43$ & $4.30$ & $4.67$ & $4.51$ & $4.51$ & $4.43$ & $4.53$ & $5.35$ & $4.44$ & $4.97$ & $4.79$ & $4.95$ & $4.53$ & $4.81$ \\
\midrule
\rowcolor{blue!8}
\cellcolor{white} & \textbf{Qwen3-14B} & $\mathbf{5.74}$ & $\mathbf{5.72}$ & $\mathbf{4.97}$ & $\mathbf{4.88}$ & $\mathbf{4.94}$ & $\mathbf{4.60}$ & $\mathbf{4.85}$ & $\mathbf{4.99}$ & $\mathbf{4.96}$ & $\mathbf{4.98}$ & $\mathbf{4.95}$ & $\mathbf{4.97}$ & $\mathbf{4.61}$ & $\mathbf{4.88}$ & $\mathbf{6.25}$ & $\mathbf{6.28}$ & $\mathbf{4.89}$ & $\mathbf{4.86}$ & $\mathbf{4.75}$ & $\mathbf{4.70}$ & $\mathbf{4.80}$ \\ \addlinespace
\rowcolor{blue!8}
\cellcolor{white}\multirow{-2}{*}{Ours} & \textbf{GPT-OSS-20B} & $\mathbf{7.29}$ & $\mathbf{7.10}$ & $\mathbf{4.88}$ & $\mathbf{4.82}$ & $\mathbf{4.79}$ & $\mathbf{4.65}$ & $\mathbf{4.79}$ & $\mathbf{7.15}$ & $\mathbf{6.36}$ & $\mathbf{4.58}$ & $\mathbf{4.51}$ & $\mathbf{4.59}$ & $\mathbf{4.39}$ & $\mathbf{4.52}$ & $\mathbf{7.38}$ & $\mathbf{6.48}$ & $\mathbf{4.90}$ & $\mathbf{4.80}$ & $\mathbf{4.76}$ & $\mathbf{4.44}$ & $\mathbf{4.73}$ \\
\bottomrule
\end{tabular}}
\end{table}

\paragraph{Text Quality.}
Across the three families, frontier models lack a controllable IF-quality trade-off. WIP shows the over-following failure mode. Gemini-3-pro pushes TPR@$1\%$FPR to $1.000$ but its PPL$_{\mathrm{Q}}$ inflates to $67.61$, roughly $10\times$ that of other models on WIP. The average LLM-as-a-Judge score drops to $3.94$, the lowest among all evaluated models. TSP shows the opposite failure: PPL$_{\mathrm{Q}}$ stays in a normal range ($2.78$ -- $7.21$) and LLM-as-a-Judge scores remain high ($4.58$ -- $4.98$), but no model reaches AUC $0.66$, so quality is preserved only because the constraint is not encoded at all. SA is the one ICW instruction where most models look balanced. This is because SA's constraint applies only at sentence-initial positions, so satisfying it leaves token distributions between sentences intact.

\subsubsection{Evaluation of our method on \icwb}

We apply our method to two different open-source backbones with weak initial performance, Qwen3-14B and GPT-OSS-20B. The results are presented in Tables~\ref{tab:main_results} and \ref{tab:text_quality}. Qwen3-14B has weak ICW IF ability across all three families, with TPR@$1\%$FPR of $0.020$, $0.036$, and $0.245$ on TSP, WIP, and SA, respectively. GPT-OSS-20B has near-zero TPR@$1\%$FPR on TSP and WIP ($0.010$ and $0.015$) and good performance on SA ($0.985$). After training, Qwen3-14B reaches TPR@$1\%$FPR of $0.941$/$0.996$/$0.985$, increasing its average TPR@$1\%$FPR from $0.100$ to $0.974$. GPT-OSS-20B reaches $0.922$/$0.989$/$0.992$, increasing its average from $0.337$ to $0.968$. The consistent ICW IF improvements across both backbones show that our method can effectively improve ICW IF ability across different model families.

These gains are achieved while maintaining high response quality. Across both perplexity and LLM-as-a-Judge, our trained models remain comparable to the original backbones, suggesting that the substantial gains in ICW IF are achieved without a degradation in overall response quality. The advantage is particularly clear on WIP: although Gemini-3-pro achieves near-perfect ICW IF, its response quality deteriorates substantially, with PPL$_Q$/PPL$_N$ of $67.61/53.51$ and an average judge score of $3.94$. In comparison, our trained Qwen3-14B obtains PPL$_Q$/PPL$_N$ of $4.99/4.96$ and a judge score of $4.88$, while GPT-OSS-20B obtains $7.15/6.36$ and $4.52$, respectively. Current results use one instruction template per family, a different key for each sample, and queries drawn from LFQA. Appendix~\ref{appendix:generalization} reports ICW IF performance under a different instruction template, on two unseen query domains, and across held-out watermark keys.

\begin{figure}[t]
\centering
\includegraphics[width=1\linewidth]{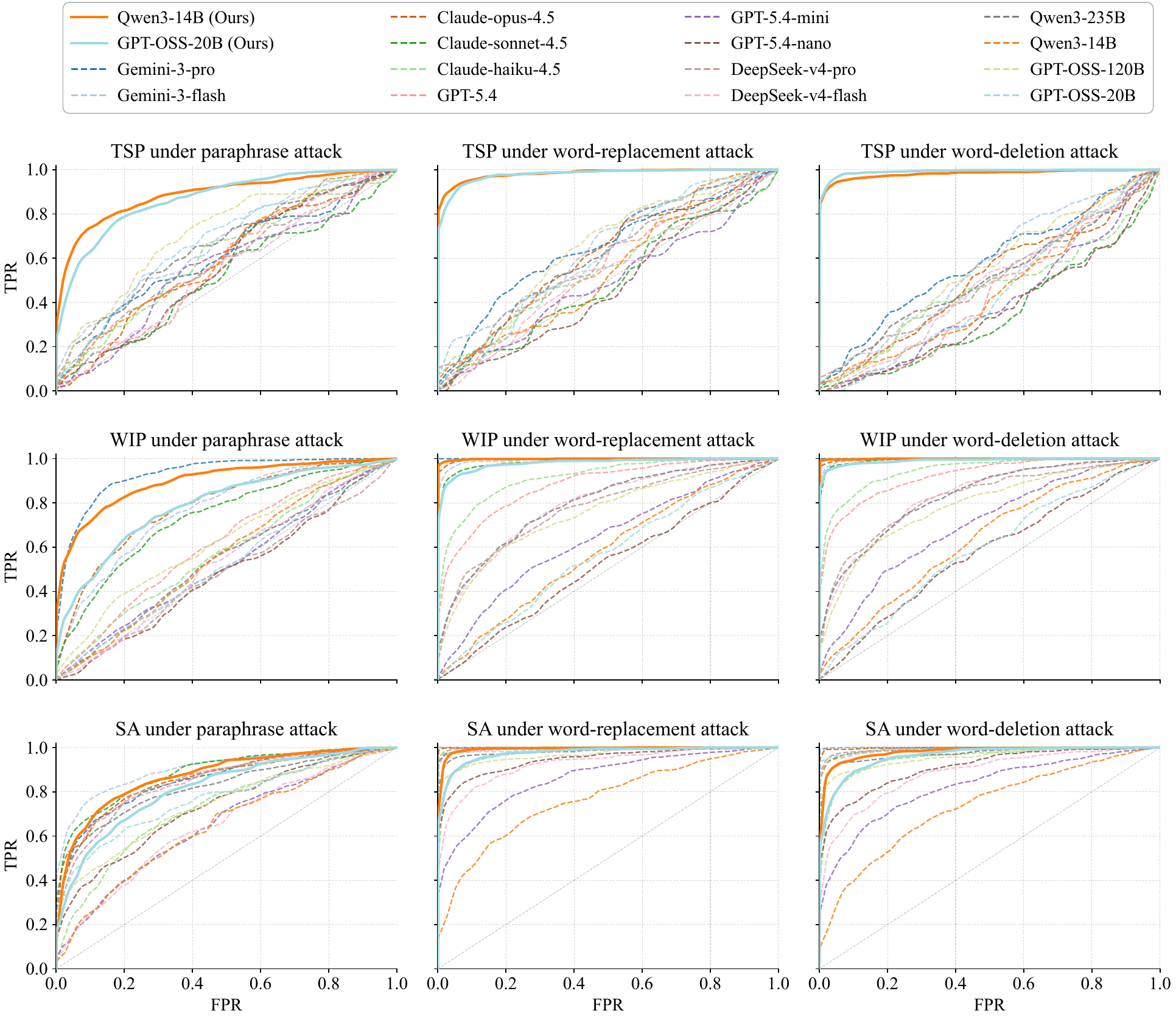}
\captionsetup{width=1\linewidth}
\caption{Robustness of ICWs under different attacks. Rows correspond to the three instruction families (TSP, WIP, SA) and columns correspond to the three attacks (paraphrase, word replacement, word deletion). Each panel overlays ROC curves of our trained Qwen3-14B and GPT-OSS-20B, and the $14$ evaluated frontier proprietary and open-source LLMs on attacked responses.}
\label{fig:robustness_roc}
\end{figure}

\subsubsection{Robustness under different attacks}

We evaluate robustness against three text perturbations applied to watermarked responses (Figure~\ref{fig:robustness_roc}). \textit{Paraphrase} rewrites the response text by an LLM. \textit{Word replacement} substitutes a fraction $30\%$ of words with synonyms. \textit{Word deletion} drops a fraction $30\%$ of words at random. Across different attacks, our trained models achieve state-of-the-art AUC on TSP. On WIP and SA, our trained models are competitive compared to the strongest baselines. Appendix~\ref{appendix:robustness} extends this evaluation to multi-round paraphrasing, back-translation, summarize-then-expand editing, and an RL-trained adaptive removal attack.

\subsection{Ablation Study}

\begin{table}[h]
\centering
\caption{Ablation of training strategies on Qwen3-14B across three ICW instruction families. PPL is the mean of the two reference-model perplexities (Qwen3-235B and Nemotron-3-Super-120B), and Judge is the mean of the four LLM-as-a-Judge dimensions. SDLP produces a strong cold start across all three families, and RL further improves IF capability. $^{*}$~marks the SA run of Base + RL, whose detection scores come with visible reward hacking in text formatting.}
\vspace{0.5em}
\label{tab:method_ablation}
\fontsize{6}{7}\selectfont
\setlength{\tabcolsep}{2.2pt}
\begin{tabular*}{\textwidth}{@{\extracolsep{\fill}}lcccccccccccc@{}}
\toprule
\multirow{2}{*}{\textbf{Models}} & \multicolumn{4}{c}{\textbf{TSP}} & \multicolumn{4}{c}{\textbf{WIP}} & \multicolumn{4}{c}{\textbf{SA}} \\
\cmidrule(lr){2-5} \cmidrule(lr){6-9} \cmidrule(lr){10-13}
& AUC $\uparrow$ & TPR@1\%FPR $\uparrow$ & PPL $\downarrow$ & Judge $\uparrow$ & AUC & TPR@1\%FPR & PPL & Judge & AUC & TPR@1\%FPR & PPL & Judge \\
\midrule
Base Model         & $0.564$ & $0.020$ & $2.99$ & $4.89$ & $0.583$ & $0.036$ & $3.53$ & $4.86$ & $0.830$ & $0.245$ & $4.13$ & $4.82$ \\ \addlinespace
Base + SFT         & $0.868$ & $0.382$ & $3.63$ & $4.89$ & $0.997$ & $0.939$ & $3.53$ & $4.70$ & $0.518$ & $0.006$ & $3.17$ & $4.88$ \\ \addlinespace
Base + RL          & $0.565$ & $0.199$ & $4.25$ & $4.75$ & $0.519$ & $0.008$ & $3.16$ & $4.79$ & $0.999^*$ & $0.994^*$ & $3.89^*$ & $4.76^*$ \\ \addlinespace
Base + SDLP        & $0.968$ & $0.822$ & $4.88$ & $4.83$ & $0.994$ & $0.925$ & $4.10$ & $4.88$ & $0.978$ & $0.849$ & $7.90$ & $4.63$ \\ \addlinespace
Base + SDLP + RL   & $0.993$ & $0.941$ & $5.73$ & $4.85$ & $0.999$ & $0.996$ & $4.98$ & $4.88$ & $0.999$ & $0.985$ & $6.27$ & $4.80$ \\
\bottomrule
\end{tabular*}
\end{table}

We ablate different training strategies against the same base model. Beyond our full method, we compare three alternatives: SFT, supervised fine-tuning as cold start; RL, GRPO with the verifier reward applied directly on the base model with no cold start; and SDLP, SDLP only without RL. Results are reported in Table~\ref{tab:method_ablation}.

\emph{SFT is instruction-dependent.} 
SFT's effectiveness varies across the three families. On WIP, SFT is effective: TPR@$1\%$FPR rises from $0.036$ to $0.939$. On TSP, the lift is partial. On SA, SFT is ineffective. SFT is therefore not a reliable cold-start strategy across ICW instructions.

\emph{RL alone is insufficient.} 
Initialized directly from the base and trained with the verifier reward, RL shows limited effectiveness on both TSP and WIP. On SA, the run numerically reaches TPR@$1\%$FPR $0.994$, but we observe obvious reward hacking in which the model introduces unnatural formatting and word-splitting artifacts to place target letters at verifier-recognized sentence-initial positions, rather than satisfying the acrostic constraint through natural sentence construction.

\emph{SDLP consistently shows strong performance on all three ICW instructions.} 
It demonstrates the effectiveness of SDLP in converting instruction-following into a decoding-time logit perturbation, which facilitates model training.

\section{Conclusion}
In this paper, we introduce \icwb, a benchmark of three verifiable ICW instructions, to extensively evaluate 14 frontier proprietary and open-source LLMs. We observe the limitations of frontier LLMs in following ICW instructions: under-following and over-following. We propose an effective, scalable, and self-contained two-stage training method (SDLP and RLVR) that makes the ICW IF of Qwen3-14B and GPT-OSS-20B achieve a better IF-quality trade-off than evaluated frontier models on TSP and WIP, and matches the strongest proprietary baselines on SA. The effectiveness of our method shows that ICWs are a promising approach to equip LLMs with a watermarking interface that third parties can invoke without access to model internals, and we hope our benchmark and training method can facilitate future research in this direction.

\clearpage

\bibliographystyle{unsrtnat}
\bibliography{references}

@article{liu2025context,
  title={In-context watermarks for large language models},
  author={Liu, Yepeng and Zhao, Xuandong and Kruegel, Christopher and Song, Dawn and Bu, Yuheng},
  journal={arXiv preprint arXiv:2505.16934},
  year={2025}
}

@article{liu2025position,
  title={Position: LLM Watermarking Should Align Stakeholders' Incentives for Practical Adoption},
  author={Liu, Yepeng and Zhao, Xuandong and Song, Dawn and Wornell, Gregory W and Bu, Yuheng},
  journal={arXiv preprint arXiv:2510.18333},
  year={2025}
}

@misc{openai2026gpt54,                                                                                                                                        
    title  = {{GPT-5.4}},                                                                                                                  
    author = {{OpenAI}},                                                                                                                                        
    year   = {2026},                                                                                                                                            
    url    = {https://developers.openai.com/api/docs/models/gpt-5.4}                           
  }

@misc{openai2026gpt54mini,                                                                                                                                        
    title  = {{GPT-5.4 mini}},                                                                                                                  
    author = {{OpenAI}},                                                                                                                                        
    year   = {2026},                                                                                                                                            
    url    = {https://developers.openai.com/api/docs/models/gpt-5.4-mini}                           
  }

@misc{openai2026gpt54nano,                                                                                                                                        
    title  = {{GPT-5.4 nano}},                                                                                                                  
    author = {{OpenAI}},                                                                                                                                        
    year   = {2026},                                                                                                                                            
    url    = {https://developers.openai.com/api/docs/models/gpt-5.4-nano}                           
  }

@misc{google2025gemini3pro,                                                                                                                                   
    title  = {{Gemini 3 Pro}},                                                                                                                       
    author = {{Google DeepMind}},                                                                                                                               
    year   = {2025},                  
    url    = {https://docs.cloud.google.com/vertex-ai/generative-ai/docs/models/gemini/3-pro}                                                                                                       
  }

@misc{google2025gemini3flash,                                                                                                                                   
    title  = {{Gemini 3 Flash}},                                                                                                                       
    author = {{Google DeepMind}},                                                                                                                               
    year   = {2025},                  
    url    = {https://docs.cloud.google.com/vertex-ai/generative-ai/docs/models/gemini/3-flash}                                                                                                       
  }

@misc{anthropic2025claudeopus45,                                                                                                                              
    title  = {{Claude Opus 4.5}},
    author = {{Anthropic}},                                                                                                                                     
    year   = {2025},                                                                                                                                            
    url    = {https://www.anthropic.com/news/claude-opus-4-5}
  }

@misc{anthropic2025claudesennet45,
    title  = {{Claude Sonnet 4.5}},
    author = {{Anthropic}},
    year   = {2025},
    url    = {https://www.anthropic.com/news/claude-sonnet-4-5}
  }

@misc{anthropic2025claudehaiku45,
    title  = {{Claude Haiku 4.5}},
    author = {{Anthropic}},
    year   = {2025},
    url    = {https://www.anthropic.com/news/claude-haiku-4-5}
  }

@techreport{deepseek2026v4,                                                                                                                                   
    title       = {{DeepSeek-V4: Towards Highly Efficient Million-Token Context Intelligence}},                                                                 
    author      = {{DeepSeek-AI}},                                                                                                                              
    year        = {2026},             
    institution = {DeepSeek},                                                                                                                                   
    url         = {https://huggingface.co/deepseek-ai/DeepSeek-V4-Pro/blob/main/DeepSeek_V4.pdf}
  }

@article{willard2023efficient,
  title={Efficient guided generation for large language models},
  author={Willard, Brandon T and Louf, R{\'e}mi},
  journal={arXiv preprint arXiv:2307.09702},
  year={2023}
}

@misc{icml2026icw,
  author={ICML 2026},   
  title={On Violations of LLM Review Policies},                  
  year={2026},  
  howpublished={\url{https://blog.icml.cc/2026/03/18/on-violations-of-llm-review-policies/}},    
  note={Accessed: 2026-03-18}                    
}

@article{zheng2023judging,
  title={Judging llm-as-a-judge with mt-bench and chatbot arena},
  author={Zheng, Lianmin and Chiang, Wei-Lin and Sheng, Ying and Zhuang, Siyuan and Wu, Zhanghao and Zhuang, Yonghao and Lin, Zi and Li, Zhuohan and Li, Dacheng and Xing, Eric and others},
  journal={Advances in neural information processing systems},
  volume={36},
  pages={46595--46623},
  year={2023}
}

@article{zheng2024sglang,
  title={Sglang: Efficient execution of structured language model programs},
  author={Zheng, Lianmin and Yin, Liangsheng and Xie, Zhiqiang and Sun, Chuyue and Huang, Jeff and Yu, Cody H and Cao, Shiyi and Kozyrakis, Christos and Stoica, Ion and Gonzalez, Joseph E and others},
  journal={Advances in neural information processing systems},
  volume={37},
  pages={62557--62583},
  year={2024}
}

@article{yao2022react,
  title={React: Synergizing reasoning and acting in language models},
  author={Yao, Shunyu and Zhao, Jeffrey and Yu, Dian and Du, Nan and Shafran, Izhak and Narasimhan, Karthik and Cao, Yuan},
  journal={arXiv preprint arXiv:2210.03629},
  year={2022}
}

@article{schick2023toolformer,
  title={Toolformer: Language models can teach themselves to use tools},
  author={Schick, Timo and Dwivedi-Yu, Jane and Dess{\`\i}, Roberto and Raileanu, Roberta and Lomeli, Maria and Hambro, Eric and Zettlemoyer, Luke and Cancedda, Nicola and Scialom, Thomas},
  journal={Advances in neural information processing systems},
  volume={36},
  pages={68539--68551},
  year={2023}
}

@article{patil2024gorilla,
  title={Gorilla: Large language model connected with massive apis},
  author={Patil, Shishir G and Zhang, Tianjun and Wang, Xin and Gonzalez, Joseph E},
  journal={Advances in Neural Information Processing Systems},
  volume={37},
  pages={126544--126565},
  year={2024}
}

@article{liu2026convexbench,
  title={ConvexBench: Can LLMs Recognize Convex Functions?},
  author={Liu, Yepeng and Huang, Yu and Wang, Yu-Xiang and Liang, Yingbin and Bu, Yuheng},
  journal={arXiv preprint arXiv:2602.01075},
  year={2026}
}

@article{mu2023can,
  title={Can LLMs Follow Simple Rules?},
  author={Mu, Norman and Chen, Sarah and Wang, Zifan and Chen, Sizhe and Karamardian, David and Aljeraisy, Lulwa and Alomair, Basel and Hendrycks, Dan and Wagner, David},
  journal={arXiv preprint arXiv:2311.04235},
  year={2023}
}

@article{zhou2023instruction,
  title={Instruction-following evaluation for large language models},
  author={Zhou, Jeffrey and Lu, Tianjian and Mishra, Swaroop and Brahma, Siddhartha and Basu, Sujoy and Luan, Yi and Zhou, Denny and Hou, Le},
  journal={arXiv preprint arXiv:2311.07911},
  year={2023}
}

@inproceedings{jiang2024followbench,
  title={Followbench: A multi-level fine-grained constraints following benchmark for large language models},
  author={Jiang, Yuxin and Wang, Yufei and Zeng, Xingshan and Zhong, Wanjun and Li, Liangyou and Mi, Fei and Shang, Lifeng and Jiang, Xin and Liu, Qun and Wang, Wei},
  booktitle={Proceedings of the 62nd Annual Meeting of the Association for Computational Linguistics (Volume 1: Long Papers)},
  pages={4667--4688},
  year={2024}
}

@inproceedings{qin2024infobench,
  title={Infobench: Evaluating instruction following ability in large language models},
  author={Qin, Yiwei and Song, Kaiqiang and Hu, Yebowen and Yao, Wenlin and Cho, Sangwoo and Wang, Xiaoyang and Wu, Xuansheng and Liu, Fei and Liu, Pengfei and Yu, Dong},
  booktitle={Findings of the Association for Computational Linguistics: ACL 2024},
  pages={13025--13048},
  year={2024}
}

@inproceedings{longpre2023flan,
  title={The flan collection: Designing data and methods for effective instruction tuning},
  author={Longpre, Shayne and Hou, Le and Vu, Tu and Webson, Albert and Chung, Hyung Won and Tay, Yi and Zhou, Denny and Le, Quoc V and Zoph, Barret and Wei, Jason and others},
  booktitle={International conference on machine learning},
  pages={22631--22648},
  year={2023},
  organization={PMLR}
}

@article{sanh2021multitask,
  title={Multitask prompted training enables zero-shot task generalization},
  author={Sanh, Victor and Webson, Albert and Raffel, Colin and Bach, Stephen H and Sutawika, Lintang and Alyafeai, Zaid and Chaffin, Antoine and Stiegler, Arnaud and Scao, Teven Le and Raja, Arun and others},
  journal={arXiv preprint arXiv:2110.08207},
  year={2021}
}

@inproceedings{wang2023self,
  title={Self-instruct: Aligning language models with self-generated instructions},
  author={Wang, Yizhong and Kordi, Yeganeh and Mishra, Swaroop and Liu, Alisa and Smith, Noah A and Khashabi, Daniel and Hajishirzi, Hannaneh},
  booktitle={Proceedings of the 61st annual meeting of the association for computational linguistics (volume 1: long papers)},
  pages={13484--13508},
  year={2023}
}

@article{ouyang2022training,
  title={Training language models to follow instructions with human feedback},
  author={Ouyang, Long and Wu, Jeffrey and Jiang, Xu and Almeida, Diogo and Wainwright, Carroll and Mishkin, Pamela and Zhang, Chong and Agarwal, Sandhini and Slama, Katarina and Ray, Alex and others},
  journal={Advances in neural information processing systems},
  volume={35},
  pages={27730--27744},
  year={2022}
}

@article{rafailov2023direct,
  title={Direct preference optimization: Your language model is secretly a reward model},
  author={Rafailov, Rafael and Sharma, Archit and Mitchell, Eric and Manning, Christopher D and Ermon, Stefano and Finn, Chelsea},
  journal={Advances in neural information processing systems},
  volume={36},
  pages={53728--53741},
  year={2023}
}

@article{dong2024self,
  title={Self-play with execution feedback: Improving instruction-following capabilities of large language models},
  author={Dong, Guanting and Lu, Keming and Li, Chengpeng and Xia, Tingyu and Yu, Bowen and Zhou, Chang and Zhou, Jingren},
  journal={arXiv preprint arXiv:2406.13542},
  year={2024}
}

@article{rao2025detecting,
  title={Detecting LLM-generated peer reviews},
  author={Rao, Vishisht Srihari and Kumar, Aounon and Lakkaraju, Himabindu and Shah, Nihar B},
  journal={PLoS One},
  volume={20},
  number={9},
  pages={e0331871},
  year={2025},
  publisher={Public Library of Science San Francisco, CA USA}
}

@article{shao2024deepseekmath,
  title={Deepseekmath: Pushing the limits of mathematical reasoning in open language models},
  author={Shao, Zhihong and Wang, Peiyi and Zhu, Qihao and Xu, Runxin and Song, Junxiao and Bi, Xiao and Zhang, Haowei and Zhang, Mingchuan and Li, YK and Wu, Yang and others},
  journal={arXiv preprint arXiv:2402.03300},
  year={2024}
}

@misc{adam2026gptzero,
      title={GPTZero: Robust Detection of LLM-Generated Texts}, 
      author={George Alexandru Adam and Alexander Cui and Edwin Thomas and Emily Napier and Nazar Shmatko and Jacob Schnell and Jacob Junqi Tian and Alekhya Dronavalli and Edward Tian and Dongwon Lee},
      year={2026},
      eprint={2602.13042},
      archivePrefix={arXiv},
      primaryClass={cs.LG},
      url={https://arxiv.org/abs/2602.13042}, 
}

@inproceedings{mitchell2023detectgpt,
  title={Detectgpt: Zero-shot machine-generated text detection using probability curvature},
  author={Mitchell, Eric and Lee, Yoonho and Khazatsky, Alexander and Manning, Christopher D and Finn, Chelsea},
  booktitle={International conference on machine learning},
  pages={24950--24962},
  year={2023},
  organization={PMLR}
}

@article{hu2023radar,
  title={Radar: Robust ai-text detection via adversarial learning},
  author={Hu, Xiaomeng and Chen, Pin-Yu and Ho, Tsung-Yi},
  journal={Advances in neural information processing systems},
  volume={36},
  pages={15077--15095},
  year={2023}
}

@article{sadasivan2023can,
  title={Can AI-generated text be reliably detected?},
  author={Sadasivan, Vinu Sankar and Kumar, Aounon and Balasubramanian, Sriram and Wang, Wenxiao and Feizi, Soheil},
  journal={arXiv preprint arXiv:2303.11156},
  year={2023}
}

@article{liang2023gpt,
  title={GPT detectors are biased against non-native English writers},
  author={Liang, Weixin and Yuksekgonul, Mert and Mao, Yining and Wu, Eric and Zou, James},
  journal={Patterns},
  volume={4},
  number={7},
  year={2023},
  publisher={Elsevier}
}

@article{zhao2023provable,
  title={Provable robust watermarking for ai-generated text},
  author={Zhao, Xuandong and Ananth, Prabhanjan and Li, Lei and Wang, Yu-Xiang},
  journal={arXiv preprint arXiv:2306.17439},
  year={2023}
}

@article{zhao2024permute,
  title={Permute-and-Flip: An optimally stable and watermarkable decoder for LLMs},
  author={Zhao, Xuandong and Li, Lei and Wang, Yu-Xiang},
  journal={arXiv preprint arXiv:2402.05864},
  year={2024}
}

@article{he2024theoretically,
  title={Theoretically grounded framework for llm watermarking: A distribution-adaptive approach},
  author={He, Haiyun and Liu, Yepeng and Wang, Ziqiao and Mao, Yongyi and Bu, Yuheng},
  journal={arXiv preprint arXiv:2410.02890},
  year={2024}
}

@article{he2025distributional,
  title={Distributional information embedding: A framework for multi-bit watermarking},
  author={He, Haiyun and Liu, Yepeng and Wang, Ziqiao and Mao, Yongyi and Bu, Yuheng},
  journal={arXiv preprint arXiv:2501.16558},
  year={2025}
}

@article{li2025statistical,
  title={A statistical framework of watermarks for large language models: Pivot, detection efficiency and optimal rules},
  author={Li, Xiang and Ruan, Feng and Wang, Huiyuan and Long, Qi and Su, Weijie J},
  journal={The Annals of Statistics},
  volume={53},
  number={1},
  pages={322--351},
  year={2025},
  publisher={Institute of Mathematical Statistics}
}

@article{liu2024adaptive,
  title={Adaptive text watermark for large language models},
  author={Liu, Yepeng and Bu, Yuheng},
  journal={arXiv preprint arXiv:2401.13927},
  year={2024}
}

@inproceedings{kirchenbauer2023watermark,
  title={A watermark for large language models},
  author={Kirchenbauer, John and Geiping, Jonas and Wen, Yuxin and Katz, Jonathan and Miers, Ian and Goldstein, Tom},
  booktitle={International Conference on Machine Learning},
  pages={17061--17084},
  year={2023},
  organization={PMLR}
}

@misc{gumbel2023,
  title = {Watermarking of large language models},
  howpublished = {\url{https://simons.berkeley.edu/talks/scott-aaronson-ut-austin-openai-2023-08-17}},
  year = {2023},
  note = {Accessed: 2023-08},
  author={Aaronson, Scott}
}

@article{kuditipudi2023robust,
  title={Robust Distortion-free Watermarks for Language Models},
  author={Kuditipudi, Rohith and Thickstun, John and Hashimoto, Tatsunori and Liang, Percy},
  journal={arXiv preprint arXiv:2307.15593},
  year={2023}
}

@article{christ2023undetectable,
  title={Undetectable Watermarks for Language Models},
  author={Christ, Miranda and Gunn, Sam and Zamir, Or},
  journal={arXiv preprint arXiv:2306.09194},
  year={2023}
}

@article{dathathri2024scalable,
  title={Scalable watermarking for identifying large language model outputs},
  author={Dathathri, Sumanth and See, Abigail and Ghaisas, Sumedh and Huang, Po-Sen and McAdam, Rob and Welbl, Johannes and Bachani, Vandana and Kaskasoli, Alex and Stanforth, Robert and Matejovicova, Tatiana and others},
  journal={Nature},
  volume={634},
  number={8035},
  pages={818--823},
  year={2024},
  publisher={Nature Publishing Group UK London}
}

@article{liu2024survey,
  title={A survey of text watermarking in the era of large language models},
  author={Liu, Aiwei and Pan, Leyi and Lu, Yijian and Li, Jingjing and Hu, Xuming and Zhang, Xi and Wen, Lijie and King, Irwin and Xiong, Hui and Yu, Philip},
  journal={ACM Computing Surveys},
  volume={57},
  number={2},
  pages={1--36},
  year={2024},
  publisher={ACM New York, NY}
}

@article{liu2023semantic,
  title={A semantic invariant robust watermark for large language models},
  author={Liu, Aiwei and Pan, Leyi and Hu, Xuming and Meng, Shiao and Wen, Lijie},
  journal={arXiv preprint arXiv:2310.06356},
  year={2023}
}

@inproceedings{liu2023unforgeable,
  title={An unforgeable publicly verifiable watermark for large language models},
  author={Liu, Aiwei and Pan, Leyi and Hu, Xuming and Li, Shuang and Wen, Lijie and King, Irwin and Yu, Philip S},
  booktitle={The twelfth international conference on learning representations},
  year={2023}
}

@article{liu2024image,
  title={Image watermarks are removable using controllable regeneration from clean noise},
  author={Liu, Yepeng and Song, Yiren and Ci, Hai and Zhang, Yu and Wang, Haofan and Shou, Mike Zheng and Bu, Yuheng},
  journal={arXiv preprint arXiv:2410.05470},
  year={2024}
}

@article{yang2025tokenpure,
  title={TokenPure: Watermark Removal through Tokenized Appearance and Structural Guidance},
  author={Yang, Pei and Liu, Yepeng and Peng, Kelly and Gao, Yuan and Song, Yiren},
  journal={arXiv preprint arXiv:2512.01314},
  year={2025}
}

@article{an2025defending,
  title={Defending llm watermarking against spoofing attacks with contrastive representation learning},
  author={An, Li and Liu, Yujian and Liu, Yepeng and Zhang, Yang and Bu, Yuheng and Chang, Shiyu},
  journal={arXiv preprint arXiv:2504.06575},
  year={2025}
}

@article{liu2025dataset,
  title={Dataset protection via watermarked canaries in retrieval-augmented llms},
  author={Liu, Yepeng and Zhao, Xuandong and Song, Dawn and Bu, Yuheng},
  journal={arXiv preprint arXiv:2502.10673},
  year={2025}
}

@inproceedings{an2026reinforcement,
  title={A reinforcement learning framework for robust and secure llm watermarking},
  author={An, Li and Liu, Yujian and Liu, Yepeng and Bu, Yuheng and Zhang, Yang and Chang, Shiyu},
  booktitle={Proceedings of the 19th Conference of the European Chapter of the Association for Computational Linguistics (Volume 1: Long Papers)},
  pages={7181--7198},
  year={2026}
}

@article{wu2025analyzing,
  title={Analyzing and evaluating unbiased language model watermark},
  author={Wu, Yihan and Cui, Xuehao and Chen, Ruibo and Huang, Heng},
  journal={arXiv preprint arXiv:2509.24048},
  year={2025}
}

@article{chen2026more,
  title={More Haste, Less Speed: Weaker Single-Layer Watermark Improves Distortion-Free Watermark Ensembles},
  author={Chen, Ruibo and Wu, Yihan and Cui, Xuehao and Zhang, Jingqi and Huang, Heng},
  journal={arXiv preprint arXiv:2602.11793},
  year={2026}
}

@article{cui2026mc,
  title={MC $^{} 2$ Mark: Distortion-Free Multi-Bit Watermarking for Long Messages},
  author={Cui, Xuehao and Chen, Ruibo and Wu, Yihan and Huang, Heng},
  journal={arXiv preprint arXiv:2602.14030},
  year={2026}
}

@article{chen2025improved,
  title={Improved Unbiased Watermark for Large Language Models},
  author={Chen, Ruibo and Wu, Yihan and Guo, Junfeng and Huang, Heng},
  journal={arXiv preprint arXiv:2502.11268},
  year={2025}
}

@inproceedings{zhao2023protecting,
  title={Protecting language generation models via invisible watermarking},
  author={Zhao, Xuandong and Wang, Yu-Xiang and Li, Lei},
  booktitle={International Conference on Machine Learning},
  pages={42187--42199},
  year={2023},
  organization={PMLR}
}

@article{qu2024provably,
  title={Provably robust multi-bit watermarking for AI-generated text via error correction code},
  author={Qu, Wenjie and Yin, Dong and He, Zixin and Zou, Wei and Tao, Tianyang and Jia, Jinyuan and Zhang, Jiaheng},
  journal={arXiv e-prints},
  pages={arXiv--2401},
  year={2024}
}

@inproceedings{fernandez2023three,
  title={Three bricks to consolidate watermarks for large language models},
  author={Fernandez, Pierre and Chaffin, Antoine and Tit, Karim and Chappelier, Vivien and Furon, Teddy},
  booktitle={2023 IEEE International Workshop on Information Forensics and Security (WIFS)},
  pages={1--6},
  year={2023},
  organization={IEEE}
}

@article{giboulot2024watermax,
  title={WaterMax: breaking the LLM watermark detectability-robustness-quality trade-off},
  author={Giboulot, Eva and Furon, Teddy},
  journal={arXiv preprint arXiv:2403.04808},
  year={2024}
}

@article{hou2023semstamp,
  title={Semstamp: A semantic watermark with paraphrastic robustness for text generation},
  author={Hou, Abe Bohan and Zhang, Jingyu and He, Tianxing and Wang, Yichen and Chuang, Yung-Sung and Wang, Hongwei and Shen, Lingfeng and Van Durme, Benjamin and Khashabi, Daniel and Tsvetkov, Yulia},
  journal={arXiv preprint arXiv:2310.03991},
  year={2023}
}

@inproceedings{zhang2024remark,
  title={$\{$REMARK-LLM$\}$: A robust and efficient watermarking framework for generative large language models},
  author={Zhang, Ruisi and Hussain, Shehzeen Samarah and Neekhara, Paarth and Koushanfar, Farinaz},
  booktitle={33rd USENIX Security Symposium (USENIX Security 24)},
  pages={1813--1830},
  year={2024}
}

@article{zhu2024duwak,
  title={Duwak: Dual watermarks in large language models},
  author={Zhu, Chaoyi and Galjaard, Jeroen and Chen, Pin-Yu and Chen, Lydia Y},
  journal={arXiv preprint arXiv:2403.13000},
  year={2024}
}

@article{li2024robust,
  title={Robust detection of watermarks for large language models under human edits},
  author={Li, Xiang and Ruan, Feng and Wang, Huiyuan and Long, Qi and Su, Weijie J},
  journal={arXiv preprint arXiv:2411.13868},
  year={2024}
}

@article{chang2024postmark,
  title={Postmark: A robust blackbox watermark for large language models},
  author={Chang, Yapei and Krishna, Kalpesh and Houmansadr, Amir and Wieting, John and Iyyer, Mohit},
  journal={arXiv preprint arXiv:2406.14517},
  year={2024}
}

@article{yang2023survey,
  title={A survey on detection of {LLM}s-generated content},
  author={Yang, Xianjun and Pan, Liangming and Zhao, Xuandong and Chen, Haifeng and Petzold, Linda and Wang, William Yang and Cheng, Wei},
  journal={arXiv preprint arXiv:2310.15654},
  year={2023}
}

@article{zhao2024sokwatermark,
  title={SoK: Watermarking for AI-Generated Content},
  author={Zhao, Xuandong and Gunn, Sam and Christ, Miranda and Fairoze, Jaiden and Fabrega, Andres and Carlini, Nicholas and Garg, Sanjam and Hong, Sanghyun and Nasr, Milad and Tramer, Florian and others},
  journal={arXiv preprint arXiv:2411.18479},
  year={2024}
}

@inproceedings{ci2024ringid,
  title={Ringid: Rethinking tree-ring watermarking for enhanced multi-key identification},
  author={Ci, Hai and Yang, Pei and Song, Yiren and Shou, Mike Zheng},
  booktitle={European conference on computer vision},
  pages={338--354},
  year={2024},
  organization={Springer}
}

@misc{qwen3technicalreport,
      title={Qwen3 Technical Report}, 
      author={Qwen Team},
      year={2025},
      eprint={2505.09388},
      archivePrefix={arXiv},
      primaryClass={cs.CL},
      url={https://arxiv.org/abs/2505.09388}, 
}

@misc{openai2025gptoss,
      title={gpt-oss-120b \& gpt-oss-20b Model Card},
      author={OpenAI},
      year={2025},
      eprint={2508.10925},
      archivePrefix={arXiv},
      primaryClass={cs.CL},
      url={https://arxiv.org/abs/2508.10925}, 
}

@article{huang2025rlcracker,
  title={RLCracker: Evaluating the Worst-Case Vulnerability of LLM Watermarks with Adaptive RL Attacks},
  author={Huang, Hanbo and Zhang, Yiran and Zheng, Hao and Gong, Xuan and Li, Yihan and Liu, Lin and Liu, Zhuotao and Liang, Shiyu},
  journal={arXiv preprint arXiv:2509.20924},
  year={2025}
}

@misc{nvidia_nemotron_3_2025,
  title  = {NVIDIA Nemotron 3: Efficient and Open Intelligence},
  author = {{NVIDIA}},
  year   = {2025},
  url    = {https://arxiv.org/abs/2512.20856},
  note   = {White Paper}
}

@inproceedings{zhang2024watermarks,
  title={Watermarks in the Sand: Impossibility of Strong Watermarking for Generative Models},
  author={Zhang, Hanlin and Edelman, Benjamin L. and Francati, Danilo and Venturi, Daniele and Ateniese, Giuseppe and Barak, Boaz},
  booktitle={International Conference on Machine Learning (ICML)},
  year={2024}
}

@article{he2026fundamental,
  title={Fundamental Trade-Offs in Multi-Bit Watermarking of Stochastic Processes},
  author={He, Haiyun and Liu, Yepeng and Shen, Zhuoer and Wang, Ziqiao and Mao, Yongyi and Bu, Yuheng},
  journal={arXiv preprint arXiv:2605.08826},
  year={2026}
}

@article{tsatsaronis2015bioasq,
  title={An overview of the {BIOASQ} large-scale biomedical semantic indexing and question answering competition},
  author={Tsatsaronis, George and Balikas, Georgios and Malakasiotis, Prodromos and Partalas, Ioannis and Zschunke, Matthias and Alvers, Michael R and Weissenborn, Dirk and Krithara, Anastasia and Petridis, Sergios and Polychronopoulos, Dimitris and Almirantis, Yannis and Pavlopoulos, John and Baskiotis, Nicolas and Gallinari, Patrick and Arti{\`e}res, Thierry and Ngonga Ngomo, Axel-Cyrille and Heino, Norman and Gaussier, Eric and Barrio-Alvers, Liliana and Schroeder, Michael and Androutsopoulos, Ion and Paliouras, Georgios},
  journal={BMC Bioinformatics},
  volume={16},
  number={1},
  pages={138},
  year={2015}
}

@inproceedings{maia2018fiqa,
  title={{WWW}'18 Open Challenge: Financial Opinion Mining and Question Answering},
  author={Maia, Macedo and Handschuh, Siegfried and Freitas, Andr{\'e} and Davis, Brian and McDermott, Ross and Zarrouk, Manel and Balahur, Alexandra},
  booktitle={Companion Proceedings of the The Web Conference 2018 (WWW '18 Companion)},
  pages={1941--1942},
  year={2018}
}

@inproceedings{wieting2022paraphrastic,
  title={Paraphrastic Representations at Scale},
  author={Wieting, John and Gimpel, Kevin and Neubig, Graham and Berg-Kirkpatrick, Taylor},
  booktitle={Proceedings of the 2022 Conference on Empirical Methods in Natural Language Processing: System Demonstrations},
  pages={379--388},
  year={2022}
}

\clearpage


\appendix
\section{Implementation Details}
\label{appendix:impl}

This appendix gives the per-family instantiation of the verifier $V_\psi$ and the logit
perturbation $\boldsymbol{\Delta}_t(\psi; \kappa)$ used by both the cold-start data
synthesis (Section~\ref{sec:tds}) and the SDLP teacher distribution (Section~\ref{sec:sdlp}). It
then reports the statistics of the cold-start dataset $\mathcal{D}_{\text{cs}}$, and
lists the instruction templates together with the procedure that maps a seed to a key (Appendix~\ref{appendix:templates}).

\subsection{Verifier and Logit Perturbation per Instruction Family}
\label{appendix:perfamily}

Rather than using the full vocabulary, we use the English-token subset $\mathcal{V}_{\mathrm{en}}\;=\;\bigl\{v\in\mathcal{V}\;:\;v\text{ is ASCII}\bigr\}\setminus\mathcal{V}_{\mathrm{special}}$. For Qwen3, $|\mathcal{V}_{\mathrm{en}}|=88{,}492$, compared to $|\mathcal{V}|=151{,}936$. We further denote by $\mathcal{V}_{\mathrm{en}}^{\,\sqcup}\subset\mathcal{V}_{\mathrm{en}}$ the subset whose surface form is a leading space followed by an ASCII letter (i.e.\ word-initial subwords); $|\mathcal{V}_{\mathrm{en}}^{\,\sqcup}|=41{,}547$. For $v\in\mathcal{V}_{\mathrm{en}}^{\,\sqcup}$, let $\phi(v)\in\{\mathtt{A},\dots,\mathtt{Z}\}$ denote the (uppercased) first letter of $v$. We restrict to $\mathcal{V}_{\mathrm{en}}$ rather than $\mathcal{V}$ to keep the context short: TSP serializes $T$ as an explicit token list in the prompt with $|T| = \gamma\,|\mathcal{V}_{\mathrm{en}}|$, so working over $\mathcal{V}_{\mathrm{en}}$ instead of $\mathcal{V}$ significantly reduces the context overhead.

For each family, $V_\psi$ instantiates the $z$-statistic in~\eqref{eq:zstat} by specifying the triple $(X,n,p_0)$, and $\boldsymbol{\Delta}_t$ instantiates~\eqref{eq:logit-perturb}.

\subsubsection{Token-Set Preference (TSP)}
\label{appendix:tsp}

\paragraph{Parameter.} $T \subseteq \mathcal{V}_{\mathrm{en}}$, sampled by drawing a uniform random subset of size $\lfloor\gamma\,|\mathcal{V}_{\mathrm{en}}|\rfloor$, where the subset ratio $\gamma\in(0,1)$ is a fixed hyperparameter. We use $\gamma=0.2$ for evaluation and $\gamma\in\{0.1,0.2,0.3\}$ for data synthesis.

\paragraph{Verifier $V_{\mathrm{TSP}}$.} Following the unique-token construction of~\cite{kirchenbauer2023watermark}, let $\mathcal{U}(y)=\{v\in\mathcal{V}_{\mathrm{en}}\,:\,v\text{ appears in }y\}$ be the set of unique tokens appearing in $y$. We set
\begin{equation*}
X\;=\;|\mathcal{U}(y)\cap T|,\qquad n\;=\;|\mathcal{U}(y)|,\qquad p_0\;=\;\gamma.
\end{equation*}
Under the null hypothesis $H_0$ (independence of $y$ from the watermark key $T$), each unique token is in $T$ with probability $\gamma$ since $T$ is a uniform random subset of $\mathcal{V}_{\mathrm{en}}$, and $z$ in~\eqref{eq:zstat} reduces to the standard binomial $z$-test.

\paragraph{Logit perturbation.} The perturbation is independent of decoding state:
\begin{equation*}
\boldsymbol{\Delta}_t^{\mathrm{TSP}}[v]\;=\;\delta_{\mathrm{TSP}}\cdot\mathbf{1}\!\left[v\in T\right],
\end{equation*}
with strength $\delta_{\mathrm{TSP}}=3.0$ during synthesis.

\subsubsection{Word-Initial Preference (WIP)}
\label{appendix:wip}

\paragraph{Parameter.} $L\subset\{\mathtt{A},\dots,\mathtt{Z}\}$ with $|L|=13$, sampled randomly from the alphabet with a key.

\paragraph{Verifier $V_{\mathrm{WIP}}$.} Let $\mathcal{U}^{\,\sqcup}(y)=\mathcal{U}(y)\cap\mathcal{V}_{\mathrm{en}}^{\,\sqcup}$. We set
\begin{equation*}
X\;=\;\bigl|\bigl\{v\in\mathcal{U}^{\,\sqcup}(y)\,:\,\phi(v)\in L\bigr\}\bigr|,
\qquad n\;=\;\bigl|\mathcal{U}^{\,\sqcup}(y)\bigr|,
\qquad p_0\;=\;\sum_{\alpha\in L}\hat f_\alpha,
\end{equation*}
where $\hat f_\alpha\;=\;|\{v\in\mathcal{V}_{\mathrm{en}}^{\,\sqcup}\,:\,\phi(v)=\alpha\}|/|\mathcal{V}_{\mathrm{en}}^{\,\sqcup}|$ is the empirical fraction of word-initial English tokens whose first letter is $\alpha$.

\paragraph{Logit perturbation.} As with TSP, the perturbation is independent of decoding state:
\begin{equation*}
\boldsymbol{\Delta}_t^{\mathrm{WIP}}[v]\;=\;\delta_{\mathrm{WIP}}\cdot\mathbf{1}\!\left[v\in\mathcal{V}_{\mathrm{en}}^{\,\sqcup}\;\wedge\;\phi(v)\in L\right],
\end{equation*}
with strength $\delta_{\mathrm{WIP}}=3.0$ during synthesis. Equivalently, $\boldsymbol{\Delta}_t^{\mathrm{WIP}}$ uniformly adds logit perturbation for every word-initial English token whose first letter lies in the in-key letter set.

\subsubsection{Sentence Acrostic (SA)}
\label{appendix:sa}

\paragraph{Parameter.} $S=s_1 s_2\cdots s_k$, drawn by sampling each character $s_i$ uniformly from a $20$-letter pool $\mathcal{P}=\mathtt{ABCDEFGHILMNOPRSTUWY}$ (the alphabet minus $\{$J, K, Q, V, X, Z$\}$, whose word-initial frequency is below $1.5\%$). We use $|S|=18$ for evaluation; for synthesis, $|S|$ is sampled uniformly in $\{18,19,20\}$ per training example.

Unlike TSP and WIP, SA constrains first letters at sentence-initial positions, which require text-level segmentation. We define an extractor $\Phi(y)$, which maps a response text $y$ to the concatenation of the first ASCII letter at each detected sentence-initial position.

\paragraph{Verifier $V_{\mathrm{SA}}$.} The observed signal is the longest common subsequence (LCS) length
\begin{equation*}
X\;=\;|\mathrm{LCS}(S,\,\Phi(y))|.
\end{equation*}
The null distribution is empirical: let $\zeta$ range over uniform random permutations of the multiset $\Phi(y)$, define the permutation null mean and variance
\begin{equation*}
\mu_S\;=\;\mathbb{E}_\zeta\!\left[|\mathrm{LCS}(S,\zeta)|\right],
\qquad
\sigma_S^{2}\;=\;\mathrm{Var}_\zeta\!\left[|\mathrm{LCS}(S,\zeta)|\right],
\end{equation*}
and report the $z$-statistic
\begin{equation*}
z\;=\;\frac{X-\mu_S}{\sigma_S}.
\end{equation*}
This generalizes~\eqref{eq:zstat}: under the Bernoulli null used by TSP and WIP, $(np_0,\;np_0(1-p_0))=(\mu,\sigma^2)$ admits a closed form, while SA has no such closed form because LCS depends on the whole sequence rather than the count alone. We estimate $(\mu_S,\sigma_S)$ with $N=1000$ permutations of $\Phi(y)$. Using LCS rather than per-position equality makes the verifier robust to extra sentences: if $|\Phi(y)|>k$, the secret string $S$ may still match $\Phi(y)$ as a subsequence anywhere, with mismatches and excess letters not penalized.

\paragraph{Logit perturbation.} Intuitively, we treat the secret $S=s_1\cdots s_k$ as a queue of letters to emit: at each new sentence start, we bias tokens whose first letter is the head of the queue, and pop the queue either when the model successfully emits that letter or fails in three consecutive sentences. Thus, the perturbation is \emph{stateful}.

\emph{State.} Per request we keep two integer counters: (1) a target index $\tau\in\{0,\dots,k\}$, the index of the next letter of $S$ we want to emit; (2) a fail counter $f\in\{0,1,2,3\}$, the number of sentences that have failed to match the current target letter $s_{\tau}$. Both start at $0$.

\emph{When to perturb logits.} A decoding step $t$ is marked as \emph{active} for logits perturbation iff (i)~$\tau<k$ (the secret $S$ is not yet fully emitted) and (ii)~the generated text $y_{<t}$ ends a sentence which can be detected by the extractor $\Phi$.

\emph{What logit bias to apply.} Let $\phi(v)$ be the first ASCII letter of $v$. When the step is \emph{active}, we bias every token whose first letter is the current target:
\begin{equation*}
\boldsymbol{\Delta}_t^{\mathrm{SA}}[v]\;=\;\delta_{\mathrm{SA}}\cdot\mathbf{1}\!\left[\phi(v)=s_{\tau}\right]\qquad(\text{active step}),
\end{equation*}
and $\boldsymbol{\Delta}_t^{\mathrm{SA}}=\mathbf{0}$ otherwise.

\begin{table}[h]
\centering
\caption{Cold-start dataset $\mathcal{D}_{\text{cs}}$ statistics by instruction family. $z$-scores and perplexities are computed on each split independently; $H_1$ samples carry the in-context watermark, $H_0$ samples do not.}
\vspace{0.5em}
\label{tab:dcs-stats}
\scriptsize
\setlength{\tabcolsep}{1.5pt}
\begin{tabular*}{0.7\textwidth}{@{\extracolsep{\fill}}lcccccc@{}}
\toprule
& \multicolumn{2}{c}{\textbf{Size}} & \multicolumn{2}{c}{\textbf{Mean $z$}} & \multicolumn{2}{c}{\textbf{Perplexity}} \\
\cmidrule(lr){2-3} \cmidrule(lr){4-5} \cmidrule(lr){6-7}
\textbf{Family} & $\mathcal{D}_{H_1}$ & $\mathcal{D}_{H_0}$ & $\mathcal{D}_{H_1}$ & $\mathcal{D}_{H_0}$ & $\mathcal{D}_{H_1}$ & $\mathcal{D}_{H_0}$ \\
\midrule
TSP & $3{,}379$ & $1{,}000$ & $7.86$ & $-1.07$ & $4.18$ & $3.20$ \\ \addlinespace
WIP & $1{,}865$ & $\phantom{0}800$ & $9.68$ & $-2.06$ & $4.35$ & $3.20$ \\ \addlinespace
SA  & $2{,}906$ & $\phantom{0}800$ & $7.91$ & $-0.09$ & $3.86$ & $3.21$ \\
\bottomrule
\end{tabular*}
\end{table}

\subsection{Cold-Start Dataset $\mathcal{D}_{\text{cs}}$ Statistics}
\label{appendix:dcs}

This section reports the size and $z$-statistics of the cold-start dataset $\mathcal{D}_{\text{cs}}=\mathcal{D}_{H_1}\cup\mathcal{D}_{H_0}$ defined in~\eqref{eq:cs-dataset}. Each $(\psi,\kappa,q^{(i)},y^{(i)})\in\mathcal{D}_{H_1}$ is generated by sampling $y^{(i)}\sim\mathcal{M}(\,\cdot\,|\,q^{(i)})$ with the family-specific perturbation $\boldsymbol{\Delta}_t(\psi;\kappa)$ injected at decoding time, and each $(\hat q^{(i)},\hat y^{(i)})\in\mathcal{D}_{H_0}$ is generated from $\mathcal{M}$ on a clean query with no perturbation. All queries are drawn from the LFQA training split, disjoint from the $500$ evaluation queries.

Table~\ref{tab:dcs-stats} reports $|\mathcal{D}_{H_1}|$, $|\mathcal{D}_{H_0}|$, and the average verifier score on $\mathcal{D}_{H_1}$. For all three families, mean $z$ on $\mathcal{D}_{H_1}$ is well above the detection threshold at $1\%$FPR, confirming that $\mathcal{D}_{\text{cs}}$ contains a strong watermark signal.

\subsection{Instruction Templates and Key Instantiation}
\label{appendix:templates}

This section gives the prompt text of the three ICW instruction families and the procedure that maps a seed to a watermark key. For every query, the paired negative response is generated from the query alone, so a positive/negative pair differs only by the presence of the instruction.

\begin{tcolorbox}[breakable, coltext=black, colback=gray!5, colframe=black!55, fonttitle=\bfseries\small, title={TSP instruction template}]
\begin{verbatim}
<green>
{green_tokens}
</green>

Respond to the user query. Seamlessly incorporate as many tokens
from <green> as possible without compromising text quality.
\end{verbatim}
\end{tcolorbox}

\begin{tcolorbox}[breakable, coltext=black, colback=gray!5, colframe=black!55, fonttitle=\bfseries\small, title={WIP instruction template}]
\begin{verbatim}
<green>
{green_letters}
</green>

<red>
{red_letters}
</red>

Given the <green> and <red> letter lists, respond to the user query
with clarity, accuracy, informativeness, and relevance. Favor words
beginning with letters from <green> and minimize those beginning with
letters from <red>. Never reveal the <green> and <red> letter lists
in your reply.
\end{verbatim}
\end{tcolorbox}

\begin{tcolorbox}[breakable, coltext=black, colback=gray!5, colframe=black!55, fonttitle=\bfseries\small, title={SA instruction template}]
\begin{verbatim}
You will receive a SECRET STRING and a user query. Answer the query
naturally and helpfully. While doing so, structure the answer so the
first letters of your sentences can spell out the secret string in
order. Treat the acrostic as a soft guide that you follow whenever it
does not hurt the response.

Rules:
1. Track the next unmatched letter in the secret string as the target
   letter. It starts at the first letter and only advances when a
   sentence successfully starts with it.
2. Prefer to start each new sentence with the target letter. If
   starting with the target letter would clearly hurt the response
   quality, write a non-matching sentence instead. After three
   consecutive misses, drop that letter and advance to the next.
3. Once the secret string is fully consumed, continue answering
   naturally with no further letter constraints.
4. Write in plain narrative prose. Do not visually highlight first
   letters in any way.
\end{verbatim}
\end{tcolorbox}

In every family the key is a deterministic function of a seed, so an instruction instance is described by the triple (family, seed, length or ratio parameter). Table~\ref{tab:key-instantiation} states the mapping. Training and evaluation draw from disjoint seed pools, so no evaluation key is seen during training.

\begin{table}[h]
\centering
\caption{Key instantiation per instruction family. Training and evaluation use disjoint seeds.}
\label{tab:key-instantiation}
\scriptsize
\begin{tabularx}{\textwidth}{p{0.07\linewidth} p{0.13\linewidth} X p{0.20\linewidth}}
\toprule
\textbf{Family} & \textbf{Key} & \textbf{Instantiation from a seed} & \textbf{Training instances} \\
\midrule
TSP & Token set $T \subseteq \mathcal{V}_{\mathrm{en}}$ & Sample $T \subseteq \mathcal{V}_{\mathrm{en}}$ tokens uniformly from $\mathcal{V}$ with the seed, $\gamma\in(0,1)$. & One seed per example; $\gamma$ drawn from $\{0.1, 0.2, 0.3\}$. \\
\addlinespace
WIP & Letter set $L$ & Sample $|L|$ letters uniformly from the alphabet with the seed. & One seed per example; $\gamma=0.5$. \\
\addlinespace
SA & Secret string $S$ & Sample $|S|$ letters with a seeded RNG and concatenate them into a string. & One seed per example; $|S|$ drawn uniformly from $\{18,19,20\}$. \\
\bottomrule
\end{tabularx}
\end{table}

\section{Generalization of ICW Instruction Following}
\label{appendix:generalization}

Section~\ref{sec:icw-instruction} describes an ICW instance by an instruction $\pi$, a key $\kappa$, and a query $q$. Training draws different keys, uses one instruction template per family and takes queries from LFQA. We therefore vary all three at evaluation time: Appendix~\ref{appendix:keys} reports how detection varies across evaluation keys, drawn from seeds disjoint from training; Appendix~\ref{appendix:template} replaces the instruction template; and Appendix~\ref{appendix:domain} replaces the query domain. All results here use our trained Qwen3-14B; we refer to it as Ours and to the untrained Qwen3-14B as Base.

\subsection{Detectability across watermark keys}
\label{appendix:keys}

$\kappa$ and $\psi$ are two parameters used by the RL reward, since that reward is $V_\psi(\,\cdot\,;\kappa)$. Our evaluation uses keys generated from seeds that are disjoint from those used during training, ensuring that the reported performance does not reflect memorization of specific keys. Table~\ref{tab:key-variation} reports the distribution of the verifier $z$-statistic over the evaluation set. The median sits between $4.29$ and $4.98$ across the three families.

\begin{table}[h]
\centering
\caption{Distribution of the verifier $z$-statistic across watermark keys, for our trained Qwen3-14B on the evaluation queries. Each response uses a key drawn from a seed disjoint from those used in training.}
\label{tab:key-variation}
\scriptsize
\begin{tabular*}{0.85\textwidth}{@{\extracolsep{\fill}}lccccc@{}}
\toprule
\textbf{Family} & Mean $z$ & SD & $p_{5}$ & Median $z$ & $p_{95}$ \\
\midrule
TSP & $4.22$ & $1.55$ & $1.84$ & $4.29$ & $6.53$ \\ \addlinespace
WIP & $4.27$ & $1.47$ & $1.75$ & $4.32$ & $6.49$ \\ \addlinespace
SA  & $4.95$ & $1.20$ & $3.09$ & $4.98$ & $6.89$ \\
\bottomrule
\end{tabular*}
\end{table}

\subsection{Held-out instruction templates}
\label{appendix:template}

We rewrote each family's instruction into a template distinct from the one used during training, keeping the task identical. Table~\ref{tab:template-generalization} reports the result. Under the reconstructed template, our method retains high detectability across all three families, and shows stable improvement over the base model.

\begin{table}[h]
\centering
\caption{ICW IF performance under a held-out instruction template, on the LFQA evaluation queries. Ours is our trained Qwen3-14B, Base is the untrained Qwen3-14B, and $\Delta$ is Ours $-$ Base.}
\label{tab:template-generalization}
\scriptsize
\begin{tabular*}{0.85\textwidth}{@{\extracolsep{\fill}}llccc@{}}
\toprule
\textbf{Template} & & TSP AUC $\uparrow$ & WIP AUC $\uparrow$ & SA AUC $\uparrow$ \\
\midrule
\multirow{3}{*}{Original}      & Base     & $0.564$ & $0.583$ & $0.830$ \\
                               & Ours     & $0.993$ & $0.999$ & $0.999$ \\
                               & $\Delta$ & \textit{$+0.429$} & \textit{$+0.416$} & \textit{$+0.169$} \\ \addlinespace
\multirow{3}{*}{Reconstructed} & Base     & $0.463$ & $0.588$ & $0.788$ \\
                               & Ours     & $0.997$ & $0.980$ & $0.977$ \\
                               & $\Delta$ & \textit{$+0.534$} & \textit{$+0.392$} & \textit{$+0.189$} \\
\bottomrule
\end{tabular*}
\end{table}

\subsection{Unseen query domains}
\label{appendix:domain}

We evaluate on two query sets that are different from training: BioASQ \citep{tsatsaronis2015bioasq} for the biomedical domain and FiQA \citep{maia2018fiqa} for the finance domain. Table~\ref{tab:domain-generalization} reports the result. The gap between Ours and Base is preserved outside the LFQA distribution the model was trained on.

\begin{table}[h]
\centering
\caption{ICW IF performance on two unseen query domains, under the original instruction template.}
\label{tab:domain-generalization}
\scriptsize
\begin{tabular*}{0.85\textwidth}{@{\extracolsep{\fill}}llccc@{}}
\toprule
\textbf{Query set} & \textbf{Model} & TSP AUC $\uparrow$ & WIP AUC $\uparrow$ & SA AUC $\uparrow$ \\
\midrule
\multirow{2}{*}{LFQA (training domain)} & Base & $0.564$ & $0.583$ & $0.830$ \\
                                        & Ours & $0.993$ & $0.999$ & $0.999$ \\ \addlinespace
\multirow{2}{*}{BioASQ (biomedical)}    & Base & $0.585$ & $0.502$ & $0.817$ \\
                                        & Ours & $0.941$ & $0.996$ & $0.993$ \\ \addlinespace
\multirow{2}{*}{FiQA (finance)}         & Base & $0.584$ & $0.510$ & $0.842$ \\
                                        & Ours & $0.949$ & $0.999$ & $0.996$ \\
\bottomrule
\end{tabular*}
\end{table}

\section{Robustness under Stronger and Adaptive Attacks}
\label{appendix:robustness}

Figure~\ref{fig:robustness_roc} evaluates paraphrase, word replacement, and word deletion. This section adds attacks that edit the response more aggressively, and one attack trained specifically to remove watermarks. As in Appendix~\ref{appendix:generalization}, all results use our trained Qwen3-14B.

\subsection{Multi-round paraphrasing and back-translation}
\label{appendix:strong-attacks}

We apply three further edits. \textit{Multi-round paraphrasing} paraphrases the response for one, two, or three successive rounds. \textit{Back-translation} translates the response into French and back into English. \textit{Summarize-then-expand} summarizes the response and then expands the summary back to full length. Table~\ref{tab:strong-attacks} reports detection AUC.

After three rounds of paraphrasing the watermark stays detectable, at AUC $0.853$, $0.872$, and $0.796$ for TSP, WIP, and SA. Back-translation costs at most $0.065$ AUC on any family. Summarize-then-expand hurts SA the most, cutting it to $0.597$; on TSP and WIP it stays at $0.757$ and $0.887$.

\begin{table}[h]
\centering
\caption{Detection AUC of our trained Qwen3-14B under three stronger edits.}
\label{tab:strong-attacks}
\scriptsize
\begin{tabular*}{\textwidth}{@{\extracolsep{\fill}}lcccccc@{}}
\toprule
\textbf{Family} & No attack & Paraphrase $\times 1$ & Paraphrase $\times 2$ & Paraphrase $\times 3$ & Back-translation & Summarize-expand \\
\midrule
TSP & $0.993$ & $0.890$ & $0.911$ & $0.853$ & $0.932$ & $0.757$ \\ \addlinespace
WIP & $0.999$ & $0.897$ & $0.962$ & $0.872$ & $0.977$ & $0.887$ \\ \addlinespace
SA  & $0.999$ & $0.873$ & $0.850$ & $0.796$ & $0.979$ & $0.597$ \\
\bottomrule
\end{tabular*}
\end{table}

\subsection{Adaptive attack}
\label{appendix:adaptive}

We evaluate against RLCracker \citep{huang2025rlcracker}, an RL-trained removal attack effective against logit-based and sampling-based watermarks.

The attack needs watermarked text to train on. That requirement is harder to satisfy for ICWs than for conventional watermarks: a logit-based or sampling-based watermark is embedded by the model provider, so an attacker can collect many watermarked outputs from one system, whereas an ICW is introduced by a third-party user. We nonetheless grant the attacker $100$ ICW-bearing samples.

We measure two quantities. The evasion success rate (ESR) is the fraction of attacked responses that fall below the detection threshold set at $1\%$ FPR. Semantic utility is measured by P-SP \citep{wieting2022paraphrastic}, the cosine similarity between averaged subword embeddings trained on paraphrase data, and by an LLM judge that scores on a $1$--$5$ scale how much of the reference answer's substance survives (Coverage) and whether what survives is unchanged in meaning (Fidelity).

Table~\ref{tab:rlcracker} reports both together. Under the loosest constraint reported (Coverage $\ge 2$), RLCracker evades detection on more than $97\%$ of responses, but it pays for this with semantic damage: P-SP falls from about $0.96$ to about $0.78$ and Coverage from about $4.95$ to about $3.5$. Once the attacked response is required to retain content, its advantage narrows. At Coverage $\ge 4$ its ESR drops to $0.261$/$0.373$/$0.378$, and at Coverage $\ge 4.5$ to $0.011$/$0.370$/$0.073$, at which point it no longer exceeds plain LLM paraphrasing on any family. The same pattern holds under a P-SP constraint. These findings are consistent with \citet{zhang2024watermarks}, which establishes that a strong (robust) watermark is theoretically impossible when attackers are allowed to alter semantic content without restriction.

\begin{table}[h]
\centering
\caption{RLCracker \citep{huang2025rlcracker} against LLM paraphrasing on our trained Qwen3-14B. Semantic utility is averaged over attacked responses. ESR is the fraction of responses that evade the $1\%$ FPR detection threshold, restricted to responses meeting the stated utility constraint. Lower ESR is better for the defender. Coverage $5$: every substantive claim, entity, number, and qualification is present; $4$: one or two minor details missing; $3$: several substantive points missing but the main answer is still delivered; $2$: most substance gone.}
\label{tab:rlcracker}
\scriptsize
\setlength{\tabcolsep}{3.2pt}
\begin{tabular}{llcccccccccccc}
\toprule
& \multirow{2}{*}{\textbf{Attack}} & \multicolumn{3}{c}{\textbf{Semantic utility}} & \multicolumn{4}{c}{\textbf{ESR $\downarrow$ at Coverage}} & \multicolumn{4}{c}{\textbf{ESR $\downarrow$ at P-SP}} \\
\cmidrule(lr){3-5} \cmidrule(lr){6-9} \cmidrule(lr){10-13}
& & P-SP $\uparrow$ & Cov. $\uparrow$ & Fid. $\uparrow$ & $\ge 2$ & $\ge 3$ & $\ge 4$ & $\ge 4.5$ & $\ge 0.6$ & $\ge 0.7$ & $\ge 0.8$ & $\ge 0.9$ \\
\midrule
\multirow{2}{*}{TSP} & LLM paraphrase & $0.960$ & $4.945$ & $4.992$ & $0.549$ & $0.549$ & $0.546$ & $0.541$ & $0.549$ & $0.546$ & $0.546$ & $0.532$ \\
                     & RLCracker      & $0.785$ & $3.412$ & $4.239$ & $0.983$ & $0.888$ & $0.261$ & $0.011$ & $0.952$ & $0.860$ & $0.473$ & $0.022$ \\ \addlinespace
\multirow{2}{*}{WIP} & LLM paraphrase & $0.965$ & $4.957$ & $4.996$ & $0.370$ & $0.370$ & $0.370$ & $0.370$ & $0.370$ & $0.370$ & $0.370$ & $0.364$ \\
                     & RLCracker      & $0.801$ & $3.556$ & $4.338$ & $0.978$ & $0.919$ & $0.373$ & $0.370$ & $0.969$ & $0.880$ & $0.560$ & $0.039$ \\ \addlinespace
\multirow{2}{*}{SA}  & LLM paraphrase & $0.967$ & $4.944$ & $4.976$ & $0.392$ & $0.392$ & $0.392$ & $0.387$ & $0.389$ & $0.389$ & $0.389$ & $0.384$ \\
                     & RLCracker      & $0.762$ & $3.605$ & $4.235$ & $0.986$ & $0.941$ & $0.378$ & $0.073$ & $0.969$ & $0.843$ & $0.300$ & $0.003$ \\
\bottomrule
\end{tabular}
\end{table}

\end{document}